\documentclass[twocolumn,twoside]{article}
\usepackage[utf8]{inputenc}
\usepackage{graphicx}
\graphicspath{{images/}}
\usepackage{caption}
\usepackage{subcaption}
\usepackage[square,numbers]{natbib}
\usepackage{url}
\usepackage[colorlinks,citecolor=blue,urlcolor=black]{hyperref}
\usepackage{todonotes}

\title{SyntheticFur dataset for neural rendering}
\author{
    Le, Trung\\
    \texttt{trungtle@google.com}
    \and
    Poplin, Ryan\\
    \texttt{rpoplin@google.com}
    \and
    Bertsch, Fred\\
    \texttt{fredbertsch@google.com}   
    \and
    Toor, Andeep Singh\\
    \texttt{andeeptoor@google.com}
    \and
    Oh, Margaret L.\\
    \texttt{maggieoh@google.com}
    }

\begin{document}

\maketitle

\section{Abstract}

\textit{We introduce a new dataset called SyntheticFur built specifically for machine learning training. The dataset consists of ray traced synthetic fur renders with corresponding rasterized input buffers and simulation data files. We procedurally generated approximately 140,000 images and 15 simulations with Houdini. The images consist of fur groomed with different skin primitives and move with various motions in a predefined set of lighting environments. We also demonstrated how the dataset could be used with neural rendering to significantly improve fur graphics using inexpensive input buffers by training a conditional generative adversarial network with perceptual loss. We hope the availability of such high fidelity fur renders will encourage new advances with neural rendering for a variety of applications.}

\newcommand\examplewidth{.2\linewidth}

\begin{figure*}[htbp]
    \centering
    \begin{tabular}{cccccccc}
    \subfloat{\includegraphics[width=0.1\linewidth]{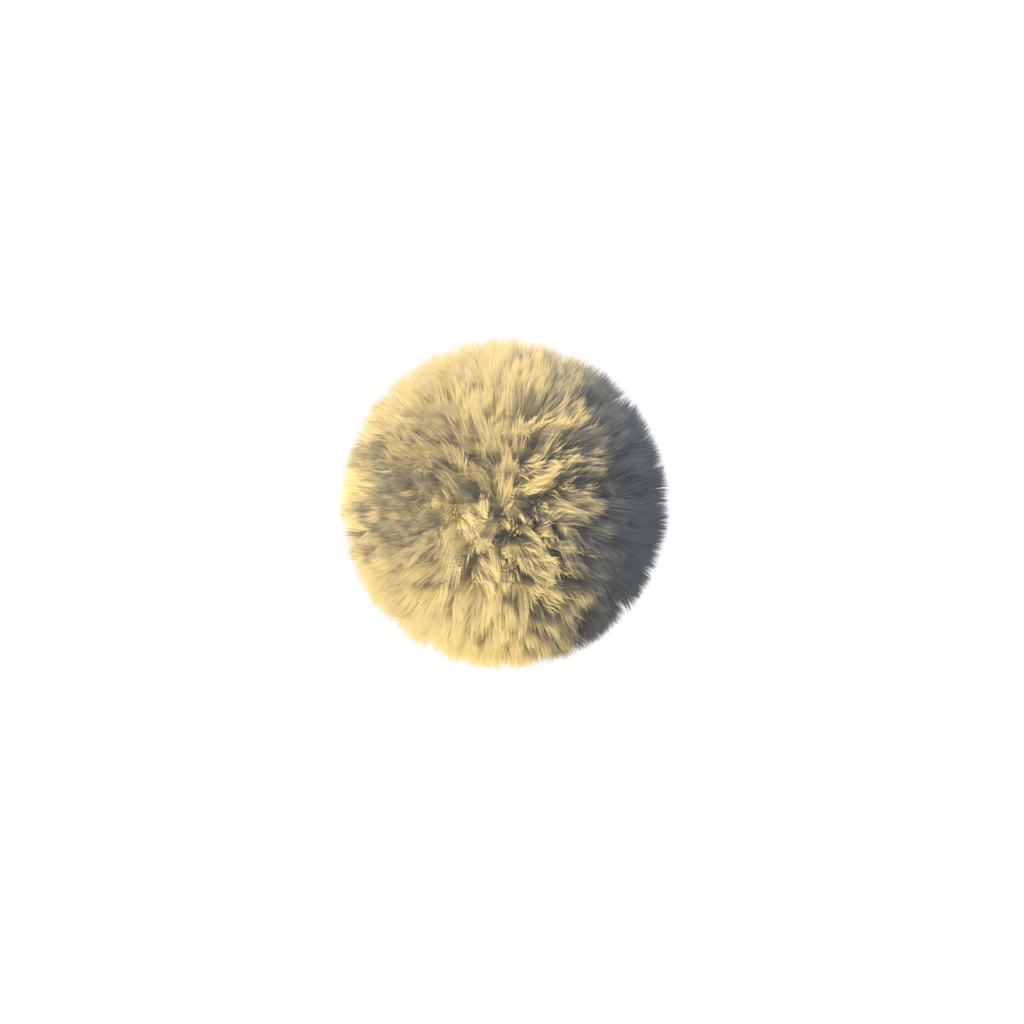}}&
    \subfloat{\includegraphics[width=0.1\linewidth]{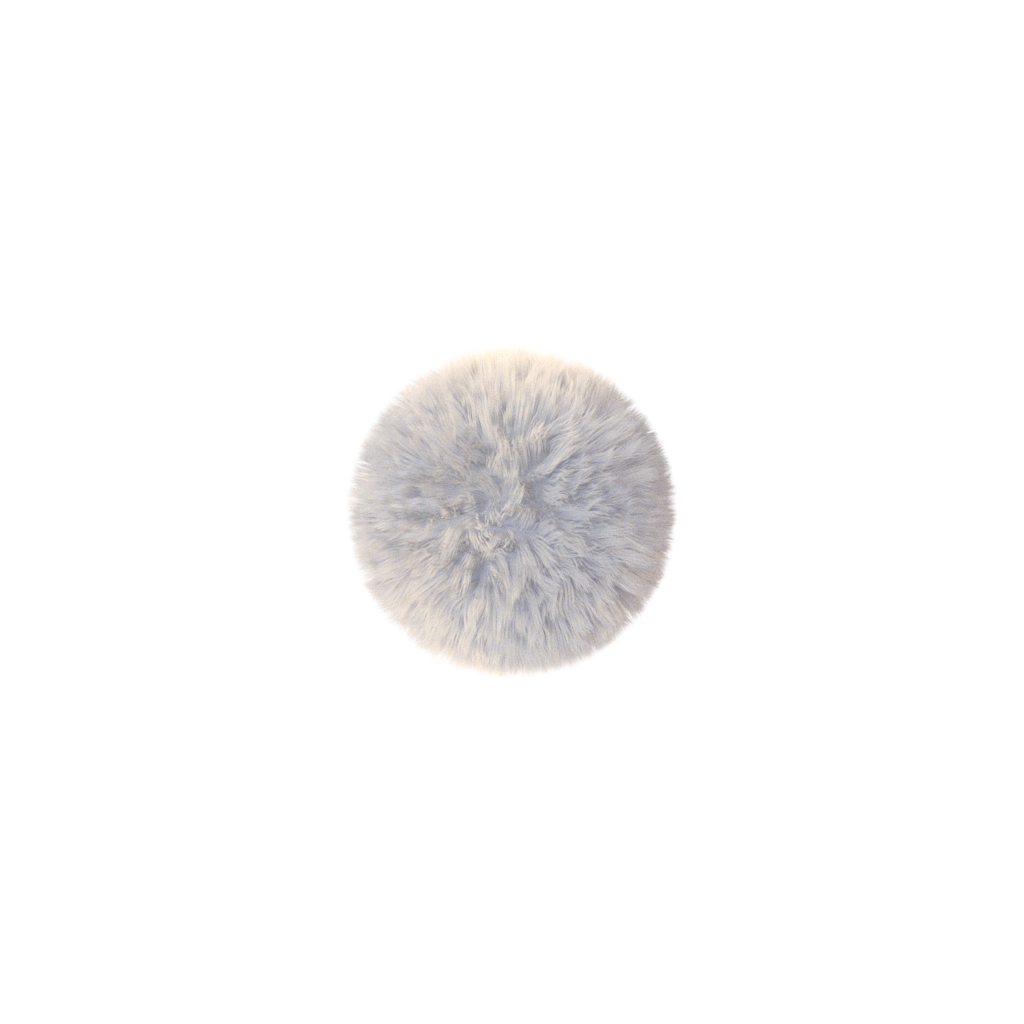}}&
    \subfloat{\includegraphics[width=0.1\linewidth]{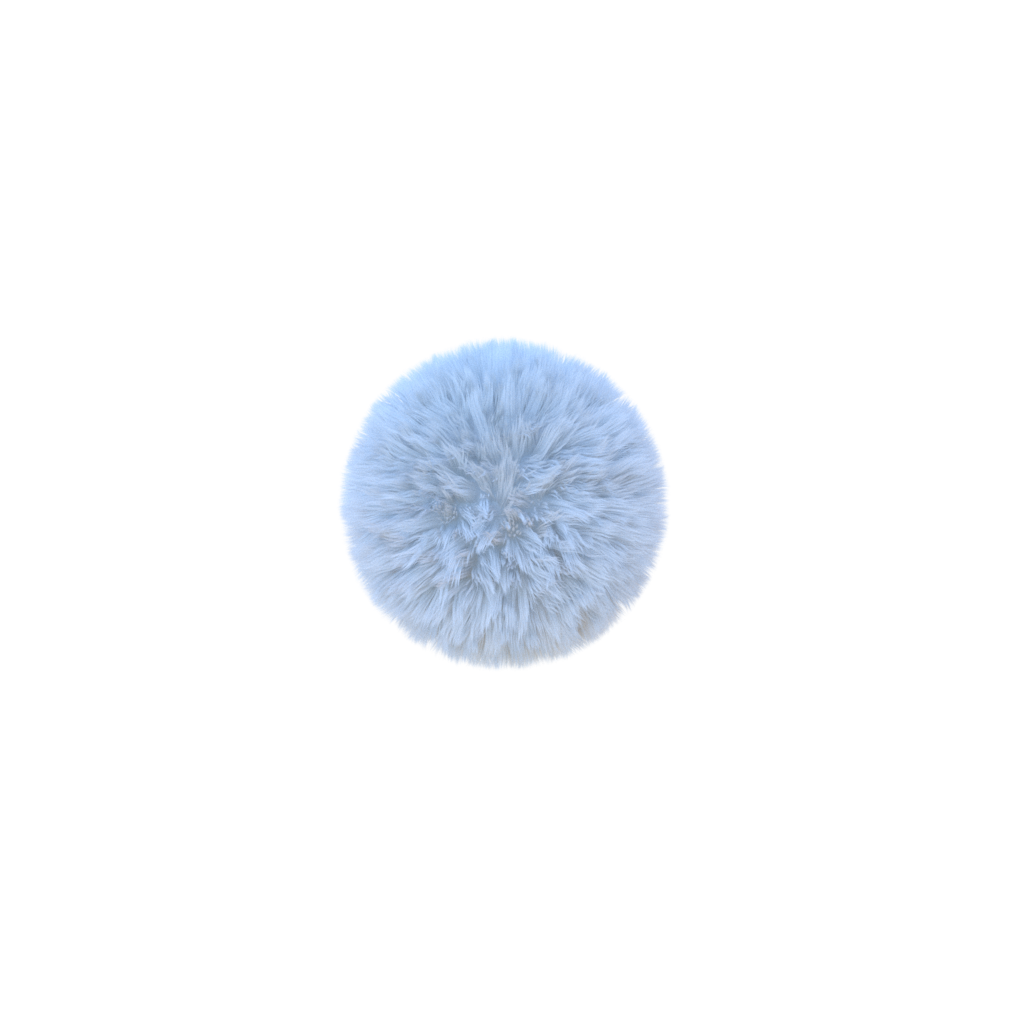}}&
    \subfloat{\includegraphics[width=0.1\linewidth]{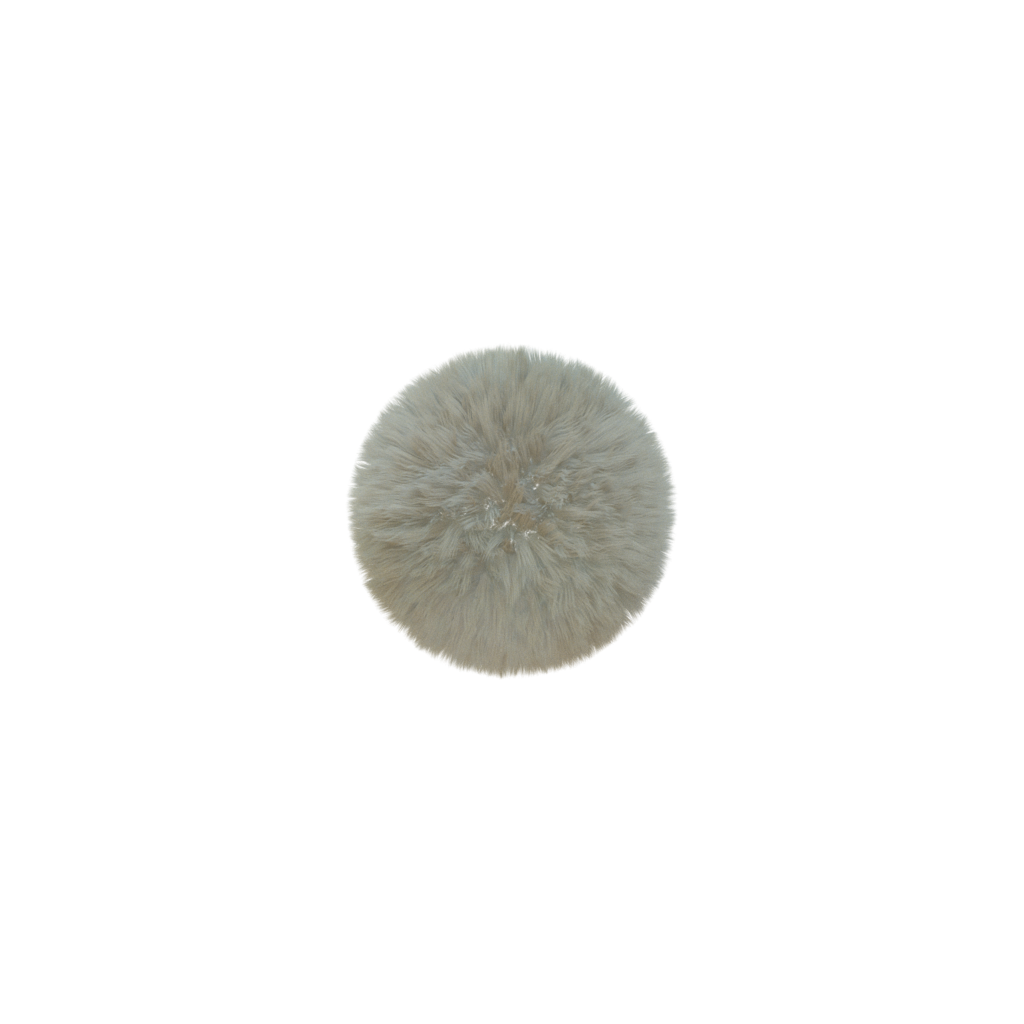}}&
    \subfloat{\includegraphics[width=0.1\linewidth]{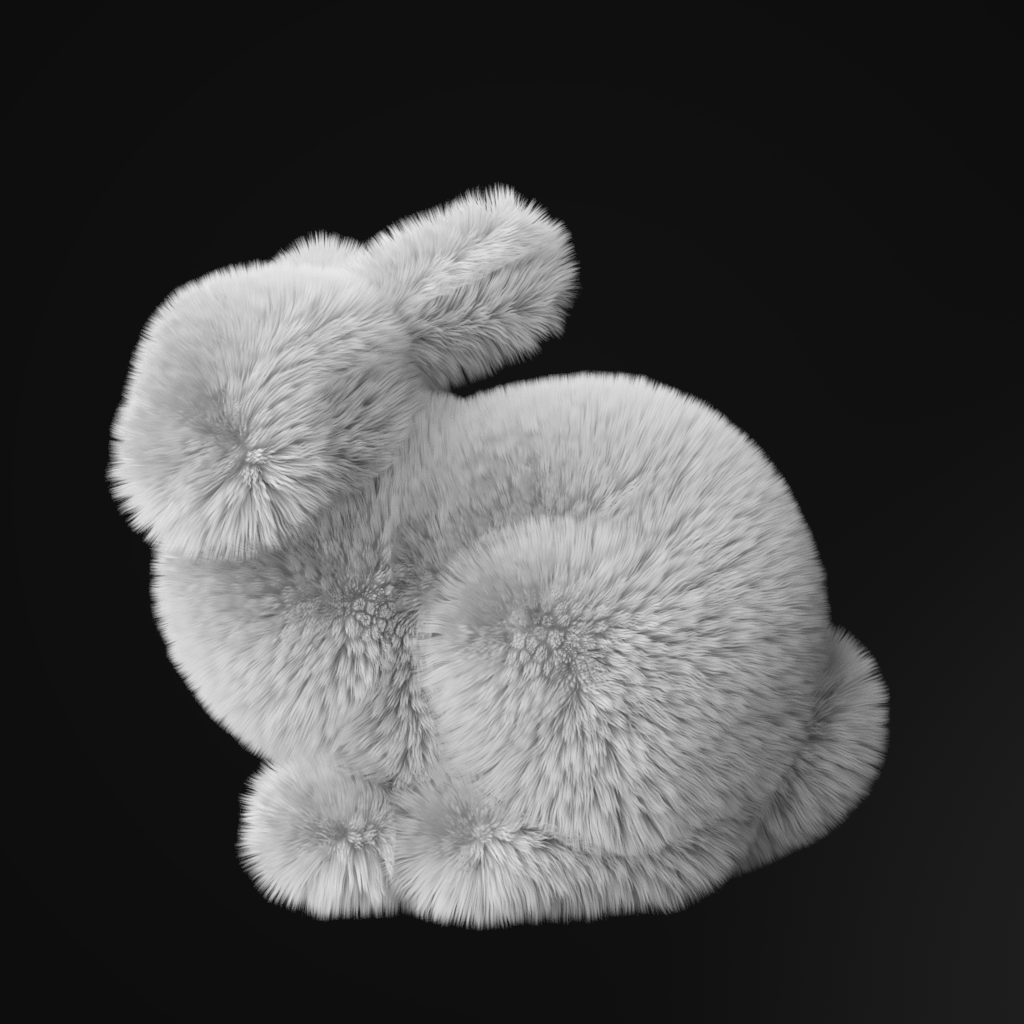}}&
    \subfloat{\includegraphics[width=0.1\linewidth]{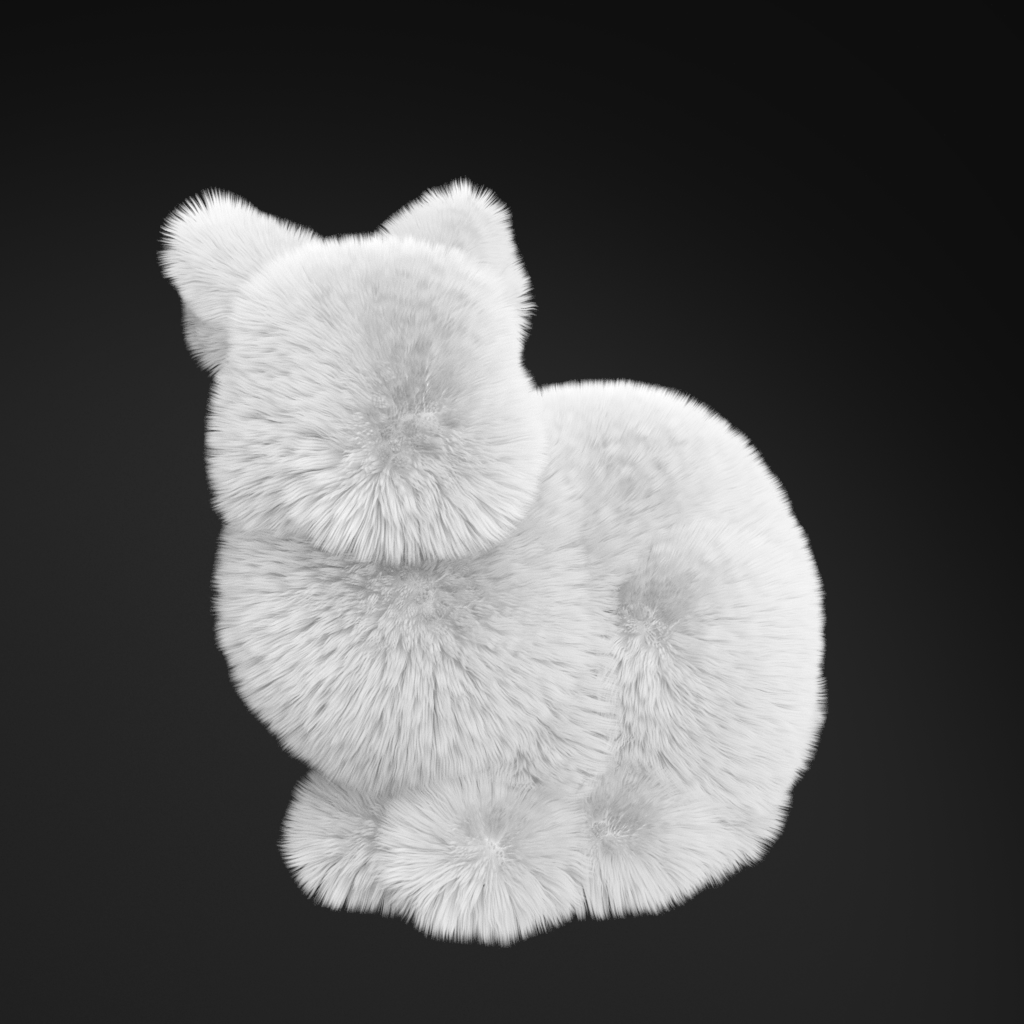}}&
    \subfloat{\includegraphics[width=0.1\linewidth]{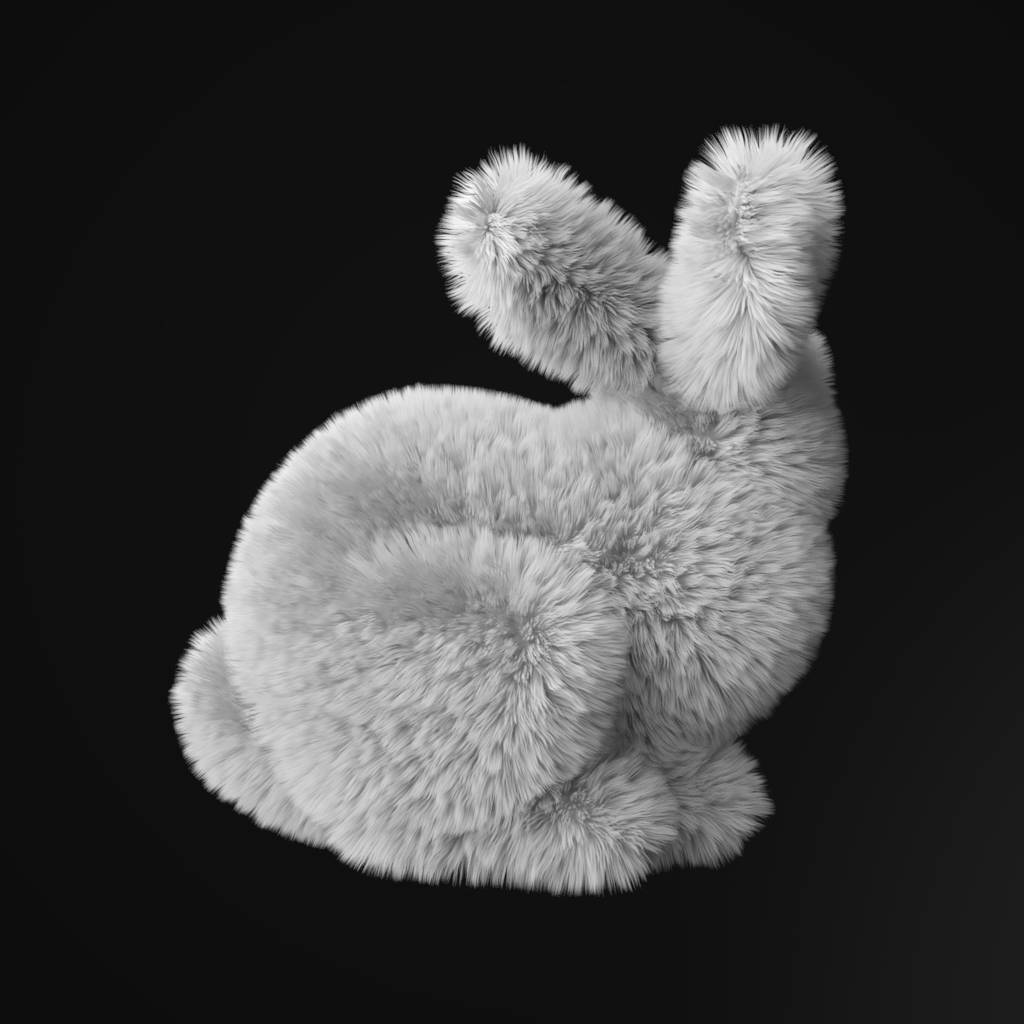}}&
    \subfloat{\includegraphics[width=0.1\linewidth]{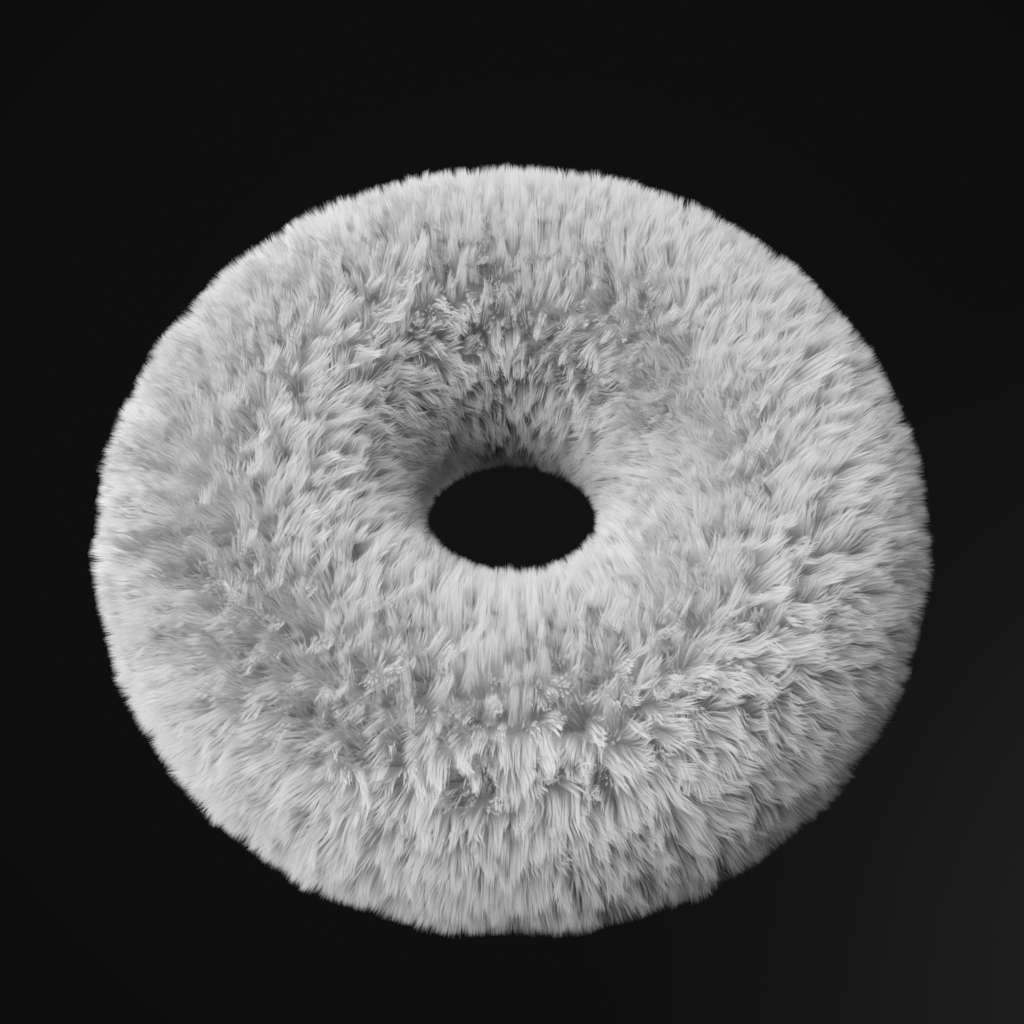}}\\
    \subfloat{\includegraphics[width=0.1\linewidth]{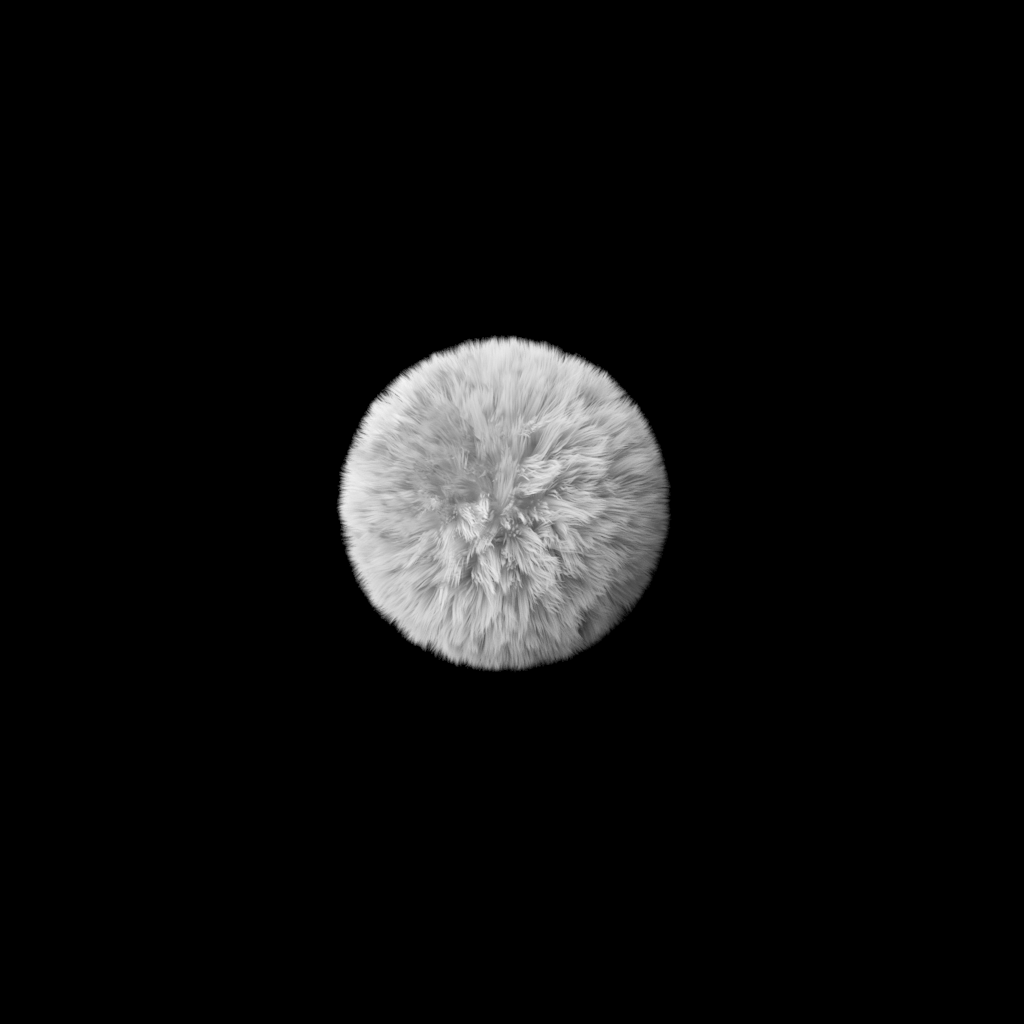}}&
    \subfloat{\includegraphics[width=0.1\linewidth]{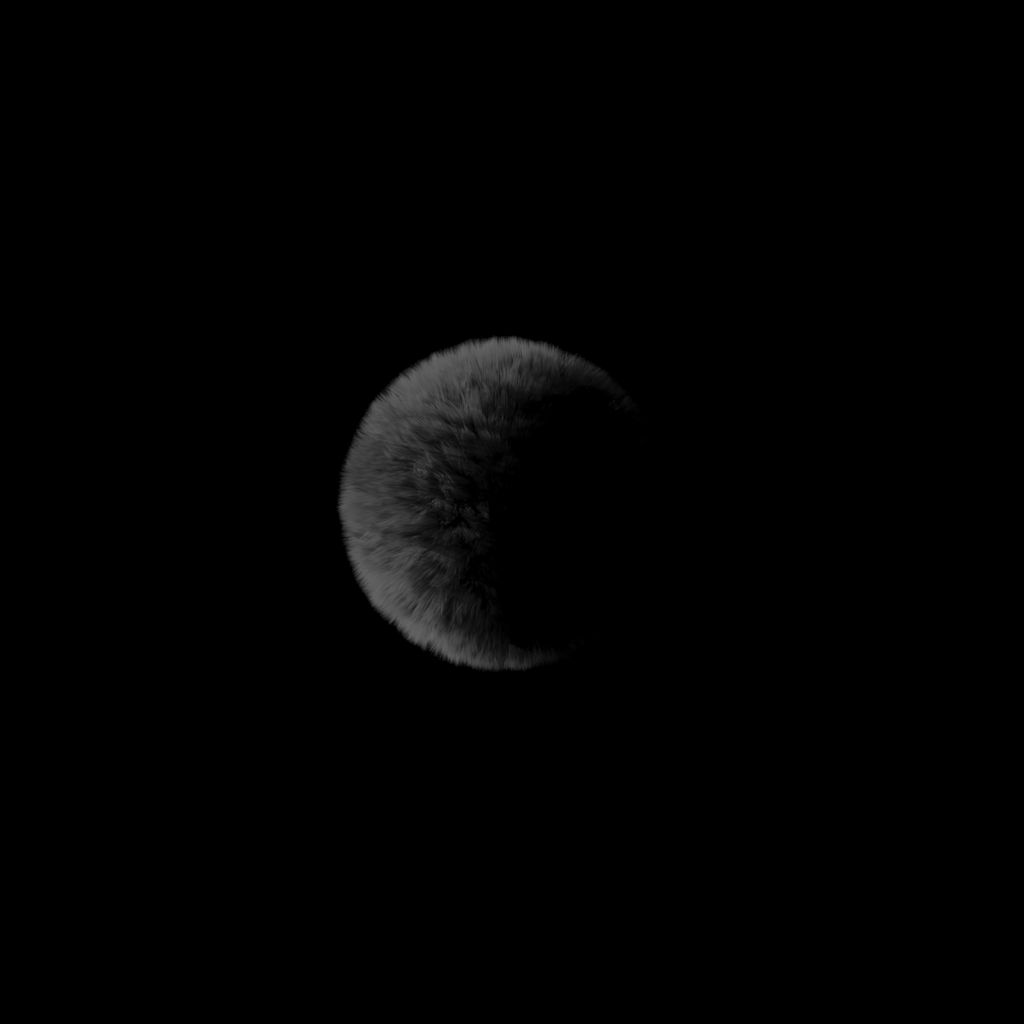}}&
    \subfloat{\includegraphics[width=0.1\linewidth]{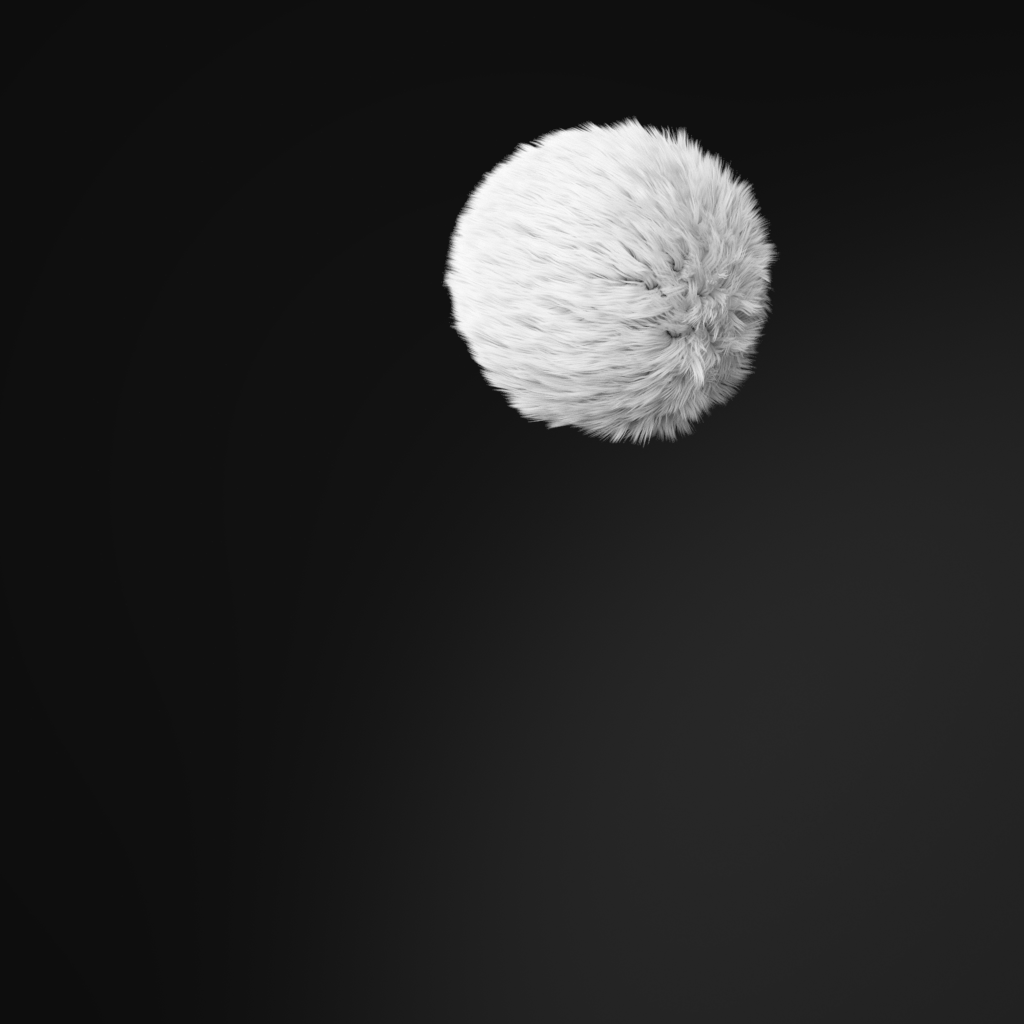}}&
    \subfloat{\includegraphics[width=0.1\linewidth]{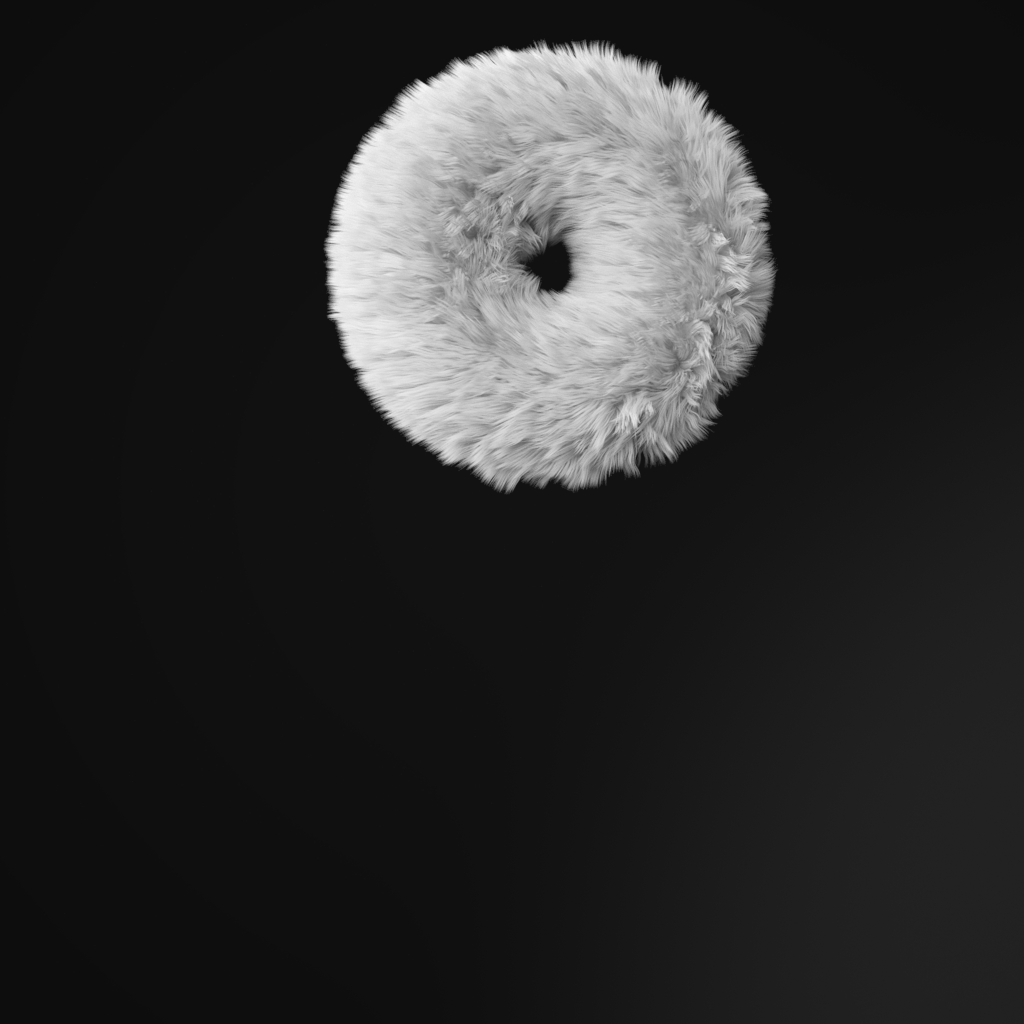}}&
    \subfloat{\includegraphics[width=0.1\linewidth]{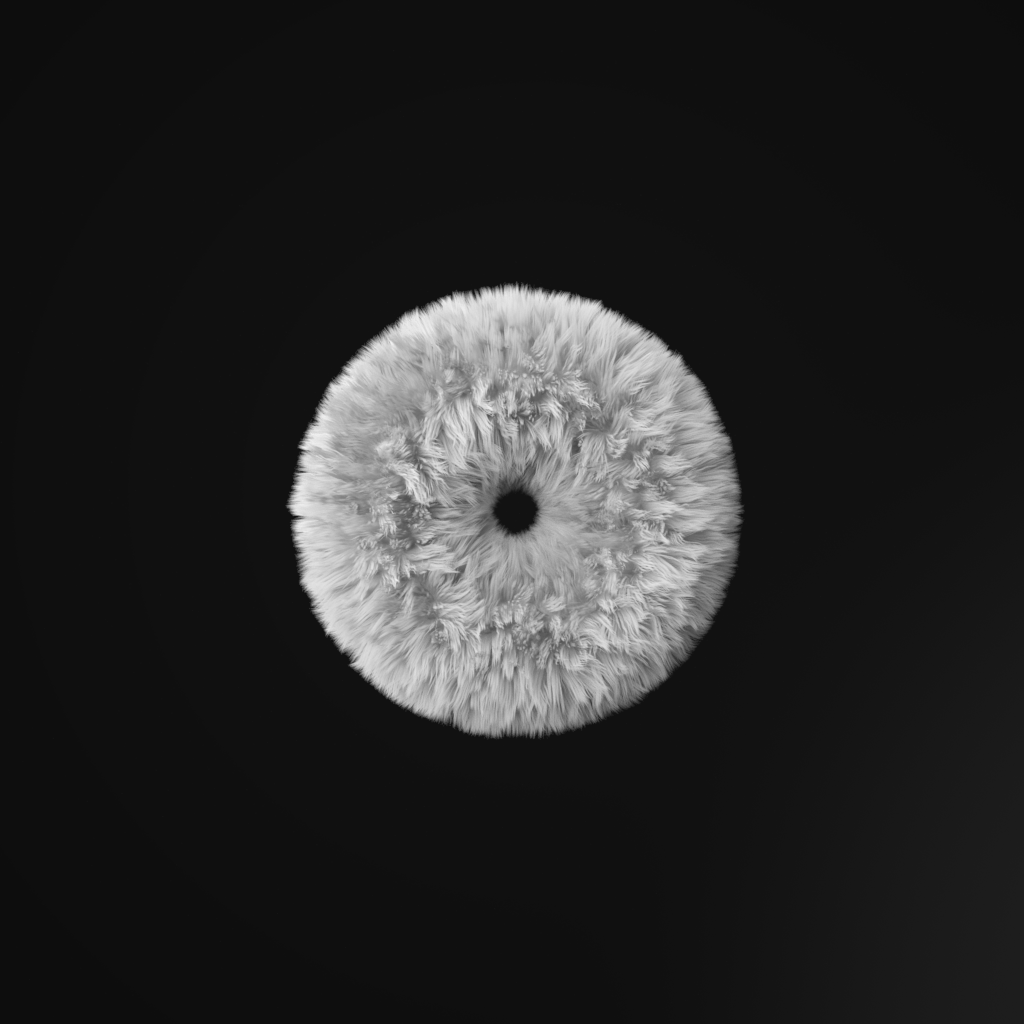}}&
    \subfloat{\includegraphics[width=0.1\linewidth]{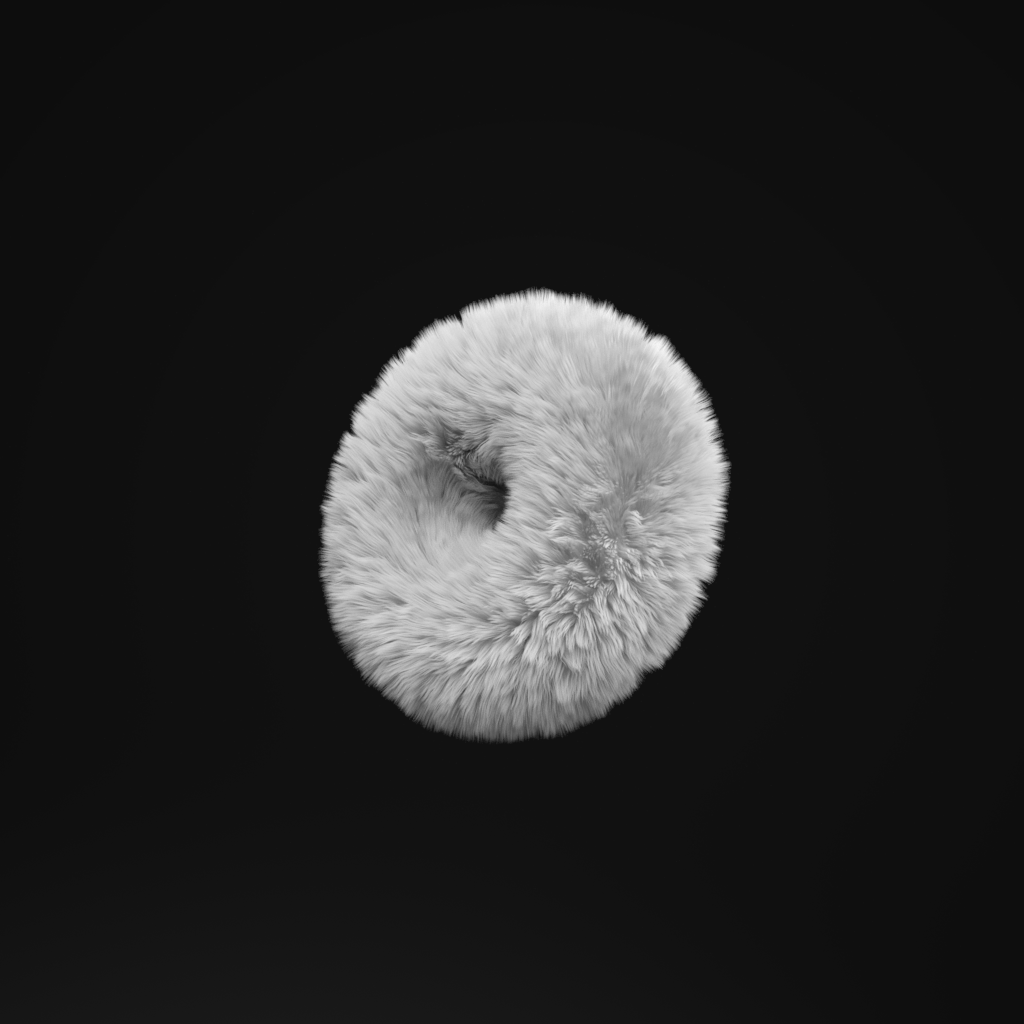}}&
    \subfloat{\includegraphics[width=0.1\linewidth]{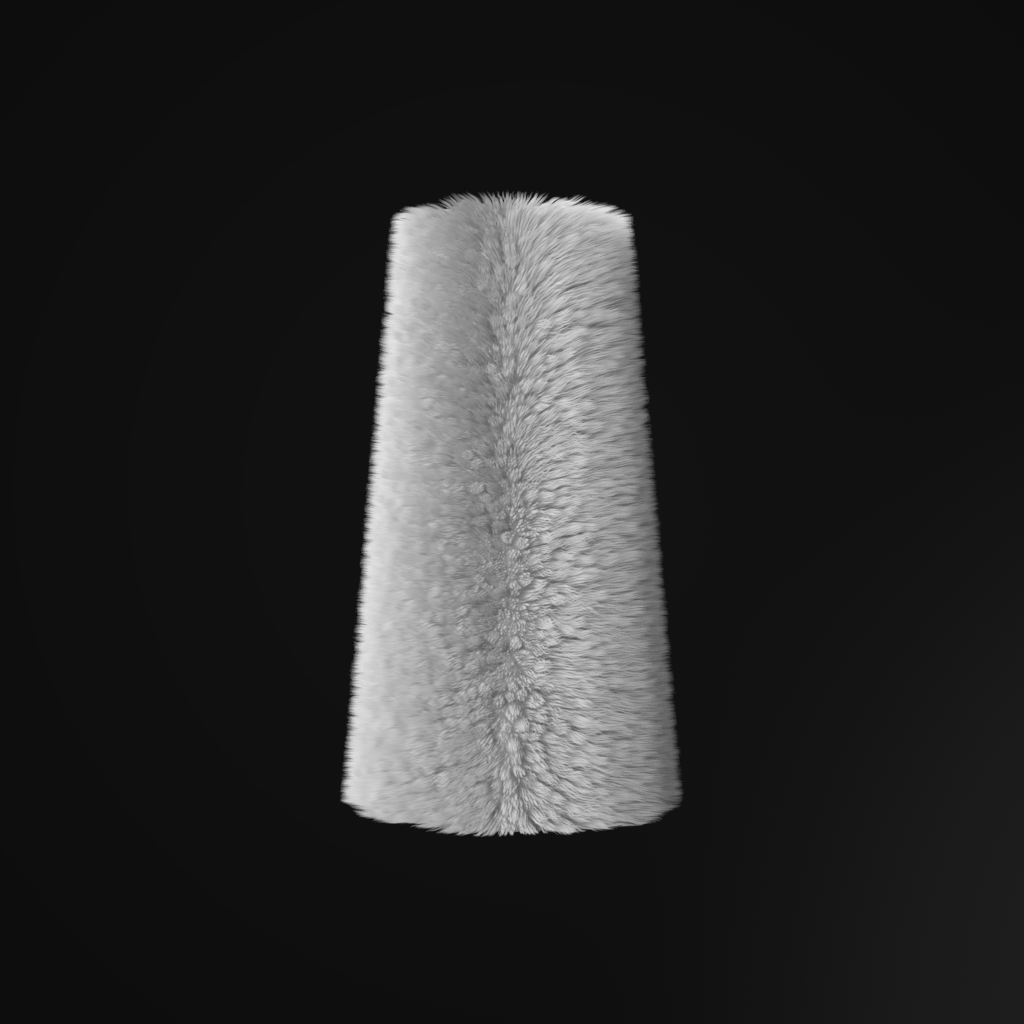}}&
    \subfloat{\includegraphics[width=0.1\linewidth]{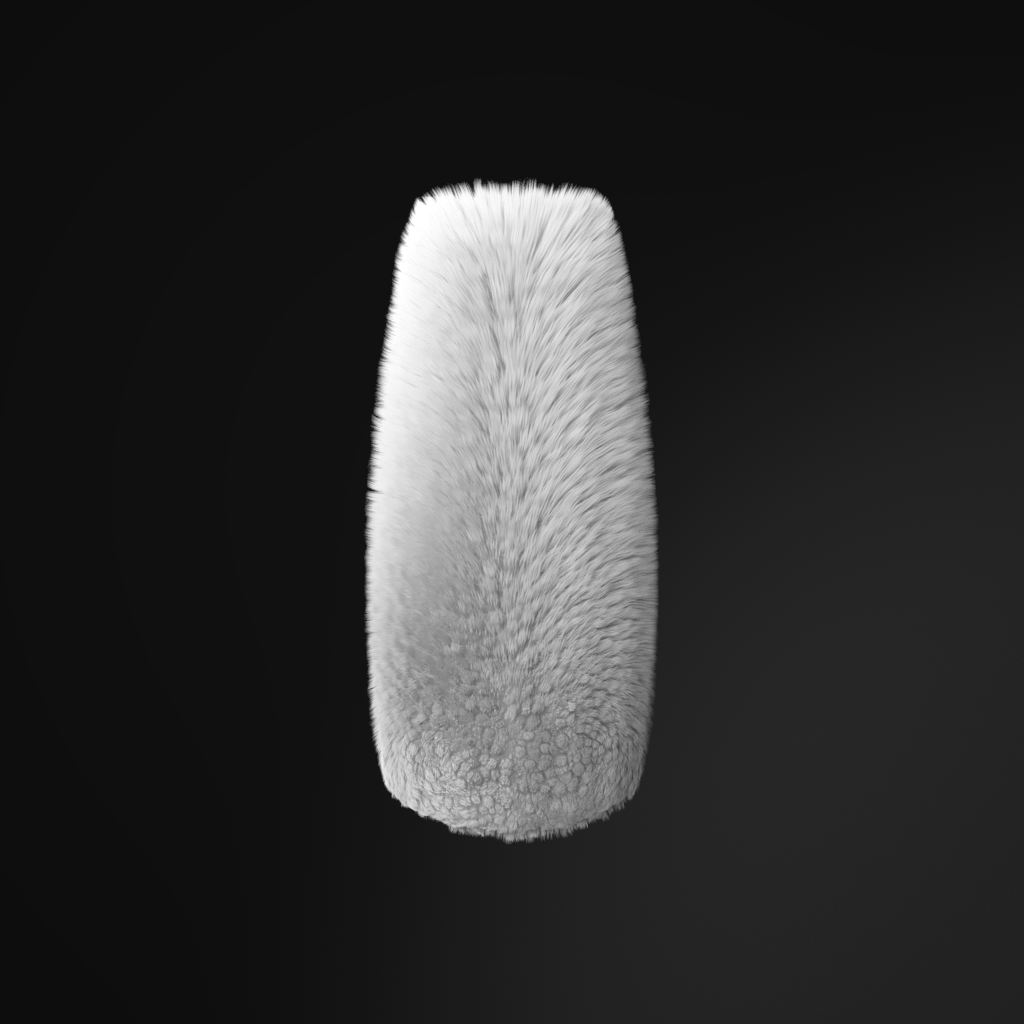}}    
    \end{tabular}
    \caption{Examples of ground truth images in SyntheticFur.}
    \label{fig:dataset examples}
\end{figure*}

\begin{figure*}[bhp]
    \centering
    \begin{subfigure}[b]{\examplewidth}
        \includegraphics[width=\linewidth]{dataset/bunny_2.png}
        \caption{Ground truth}
    \end{subfigure}
    \begin{subfigure}[b]{\examplewidth}
        \includegraphics[width=\linewidth]{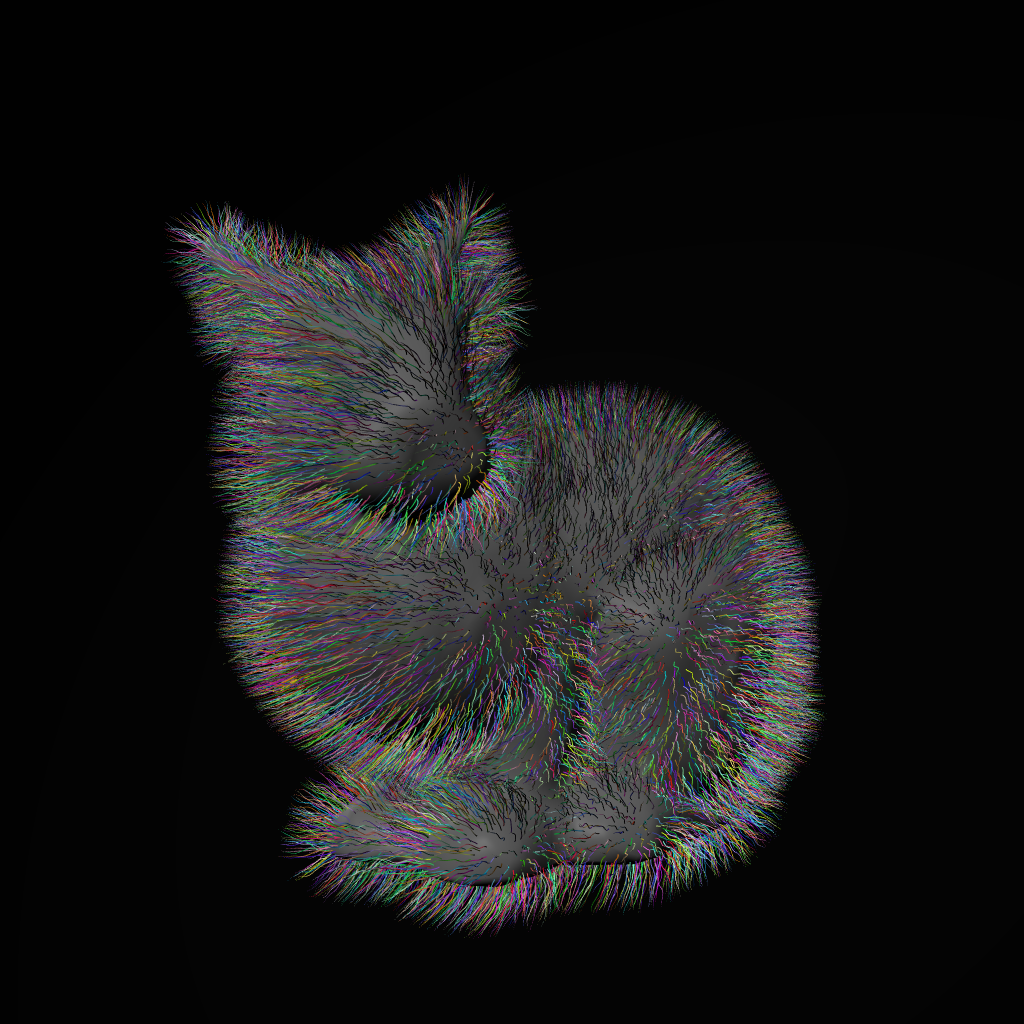}
        \caption{Colored guides}
    \end{subfigure}
    \begin{subfigure}[b]{\examplewidth}
        \includegraphics[width=\linewidth]{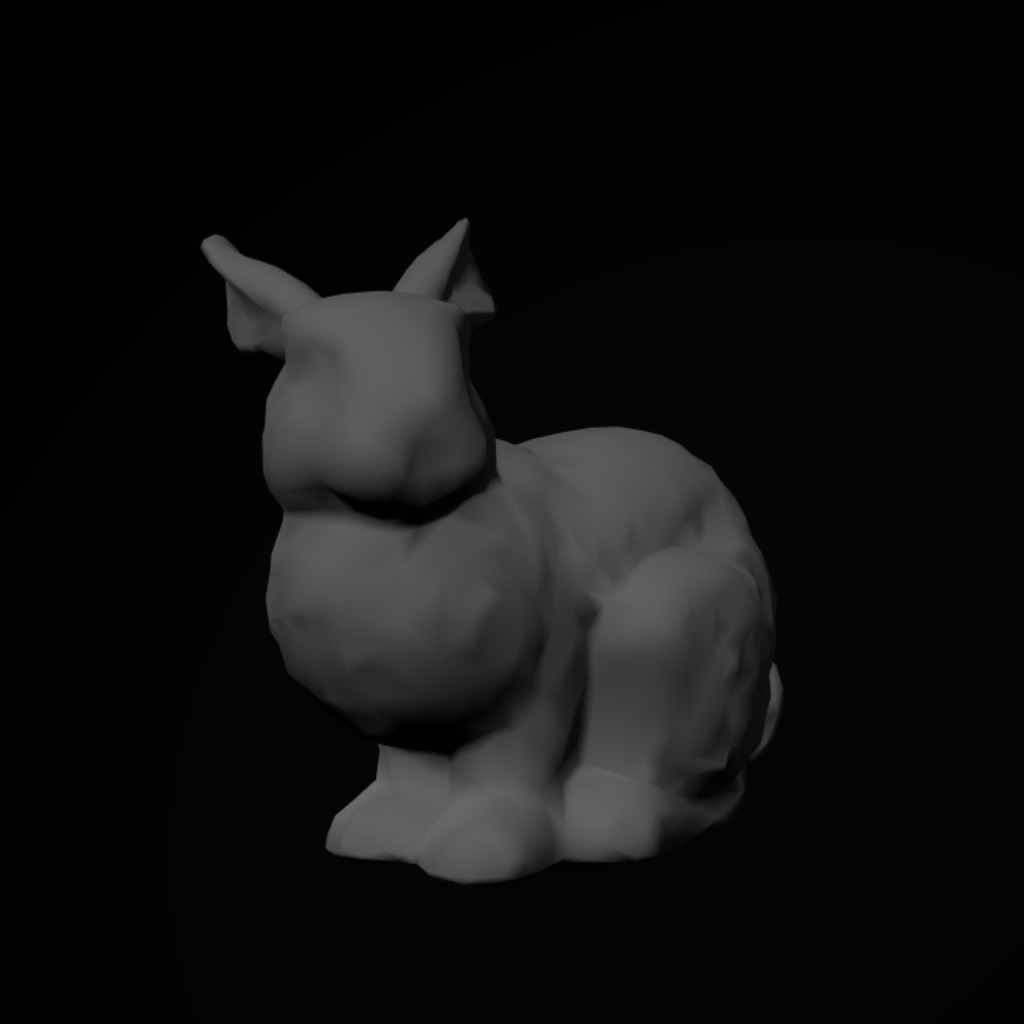}
        \caption{Lit primitive}
    \end{subfigure}
    \begin{subfigure}[b]{\examplewidth}
        \includegraphics[width=\linewidth]{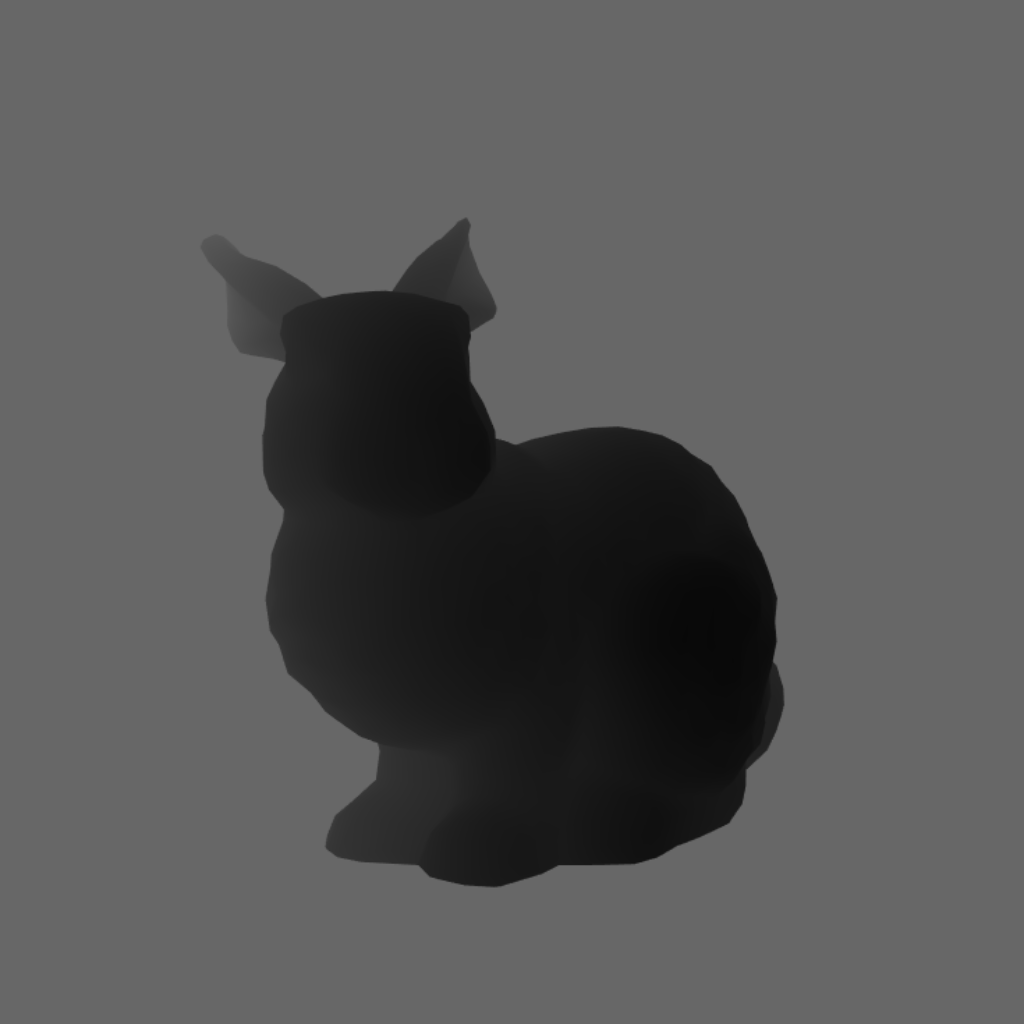}
        \caption{Scene depth}
    \end{subfigure}
    \begin{subfigure}[b]{\examplewidth}
        \includegraphics[width=\linewidth]{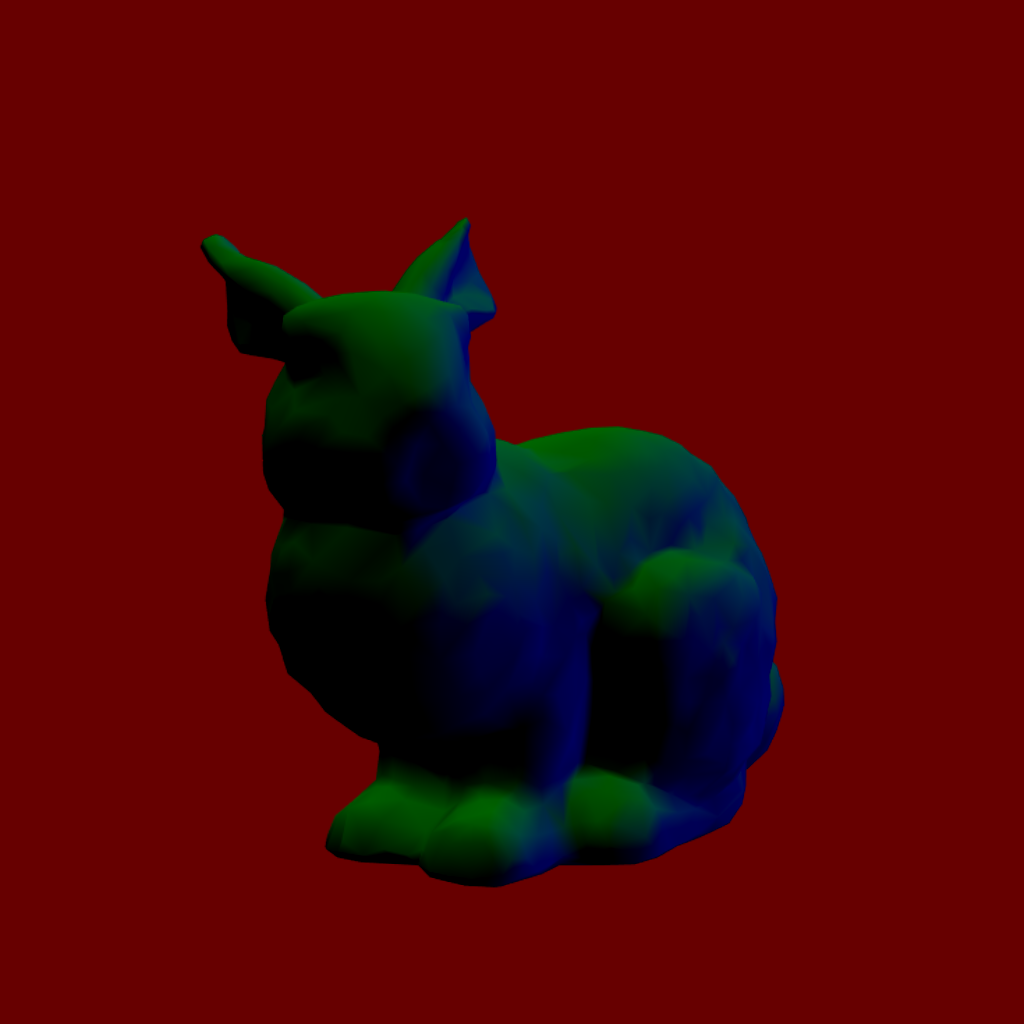}
        \caption{World normal}
    \end{subfigure}
    \begin{subfigure}[b]{\examplewidth}
        \includegraphics[width=\linewidth]{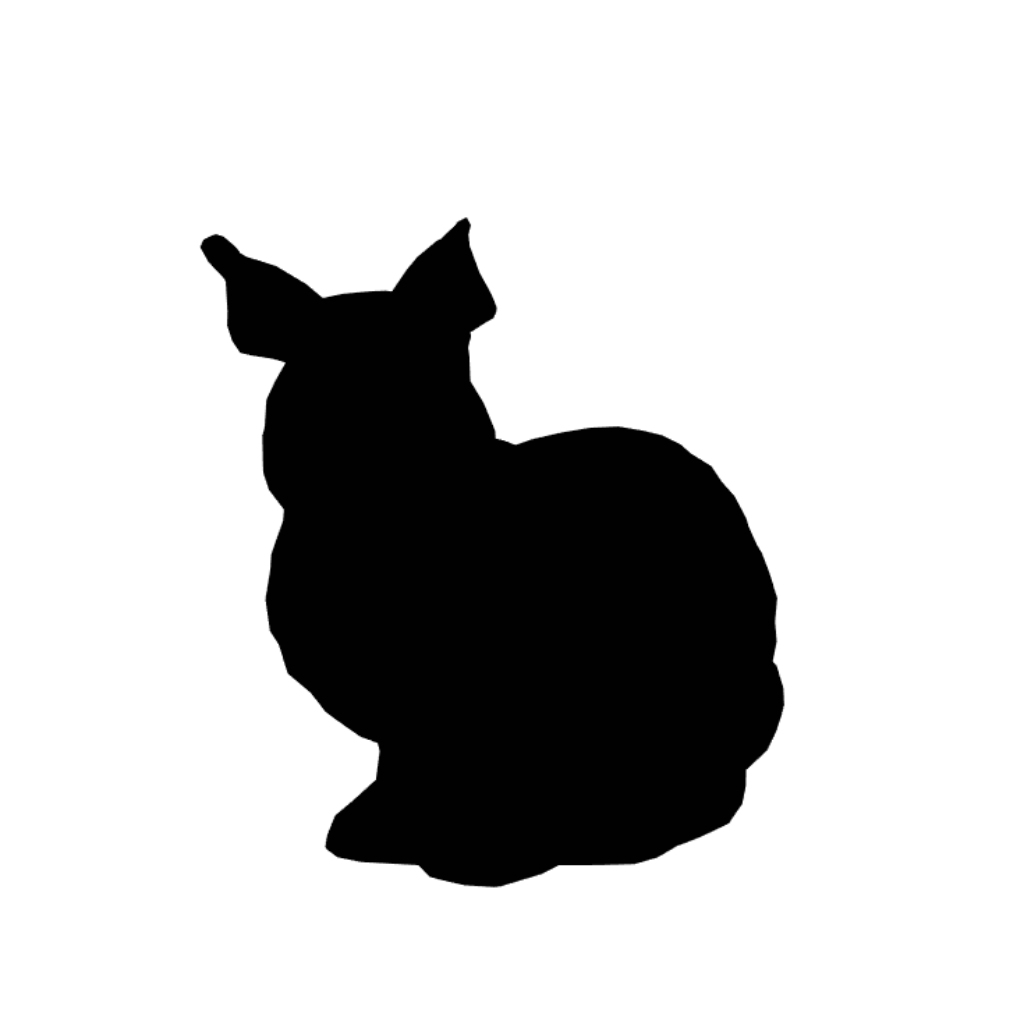}
        \caption{Mask}
    \end{subfigure}
    \begin{subfigure}[b]{\examplewidth}
        \includegraphics[width=\linewidth]{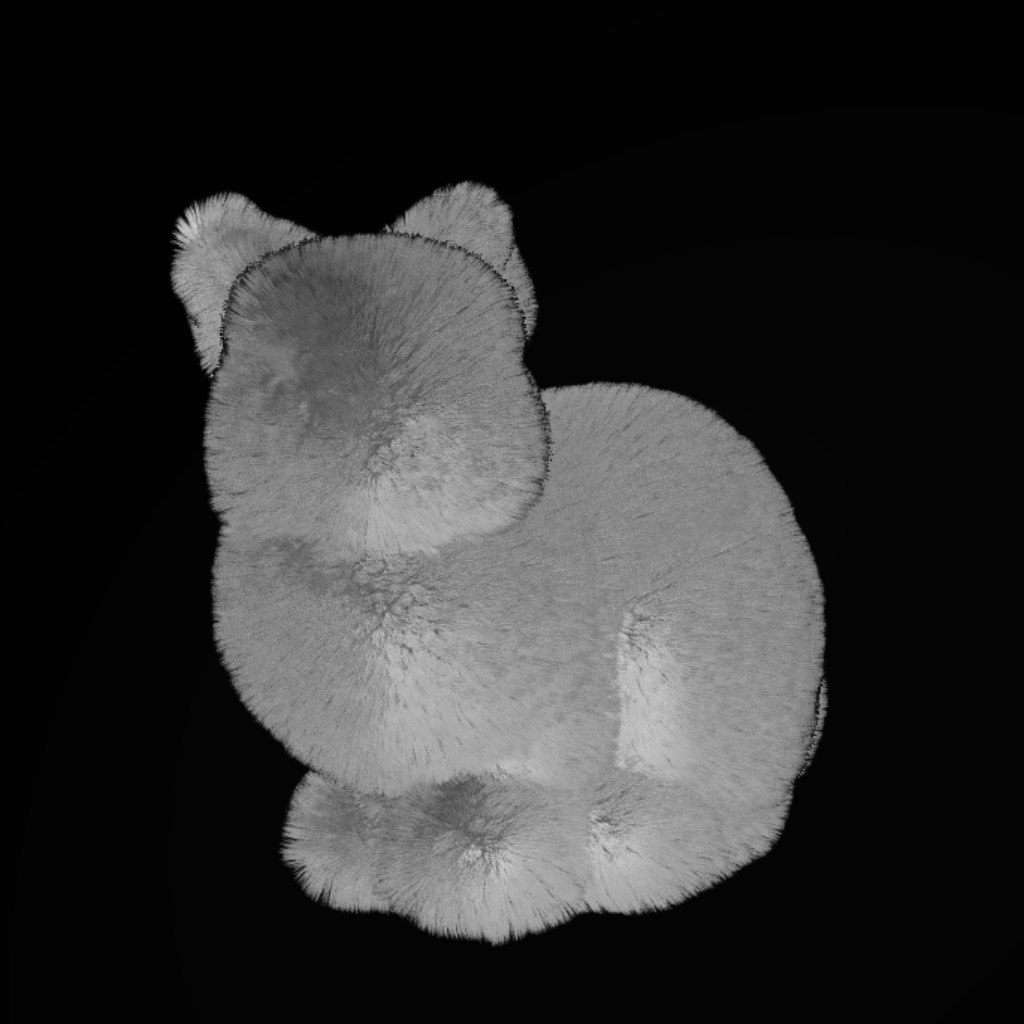}
        \caption{Rasterized}
    \end{subfigure}
    \caption{Each frame in the dataset contains a corresponding set of ground truth and input buffer images for training and evaluation.}
    \label{fig:buffer examples}
\end{figure*}
\section{Introduction}

Fur is an important element of character design in films and video games to connect the viewers to these characters with empathetic emotions. Physically based rendering of hair and fur \cite{marschner2003} \cite{kajiyakay1989} \cite{yan2017} model the hair fibers as dielectric cylinders to solve for the light rendering equation with ray tracing. Production ray tracing engines in films and animations can render fur with physically accurate visual quality and simulation \cite{productionhair}. However, real-time applications such as video games often cannot afford the same budget due to the complexity of rendering dense hair strands with transparent material, and the expensive computational cost of physics simulation. Most commonly, the solution is to pre-render the strands into hair cards and significantly reduce the number of strands to simulate. Recent advancement in game engines introduce  the new strand-based approach \cite{unrealhair} \cite{frostbitehair2019} \cite{frostbitehair2020} for real-time interactive hair that improve over hair cards. However, implementation of real-time strand-based fur rendering is still a complex task. 

The state of neural rendering \cite{tewari2020state} \cite{mildenhall2020nerf} has risen in popularity due to the increasing number of publicly available  image datasets \cite{deng2009imagenet} \cite{hasinoff2016burst} \cite{legobrick} \cite{yu2016lsun} \cite{uschairsalon} for machine learning. Many of these techniques take advantage of deep learning's effectiveness in solving image to image translation tasks. We recognize that we can bring the rendering quality of real-time fur closer to that of offline ray tracing with a similar approach and created the SyntheticFur dataset. We create the dataset specifically for deep learning from the ground up and so are able to make certain design decisions to help simplify training tasks including consistent image resolutions, backgrounds, lighting environments and conditions, fur materials, skin primitives, and simple motions. A synthetic dataset tailored for a particular training task has the advantage of being cleaner to process and simpler to analyze, with less feature engineering typically involved when working with a more generic photographic dataset.

To create the dataset, we use the Houdini software to procedurally groom skin primitives with hair strands, and render the ground truth images with ray tracing using Mantra as the rendering engine. To generate the input images, we use the Houdini viewport rasterization of the same view to closely represent common image buffers generated by a game engine. We construct several scenes of a variety of different skin primitives and motions under predefined lighting conditions.

For modeling, we use a conditional generative adversarial network (cGAN). The model takes as inputs a set of rasterized image buffers and computes GAN loss and perceptual loss against the ray traced ground truth image. We do not use the simulation data here but note that it might be a fruitful direction for future work.

In summary, our contributions are:
\begin{itemize}
  \item A high quality synthetic fur dataset with images and simulation data.
  \item An image-to-image model that uses this dataset to transform from inexpensive inputs to high quality renders.
\end{itemize}

\section{Related Work}

There are many image datasets freely available for use in computer vision and machine learning research, from more general images such as ImageNet \cite{deng2009imagenet} \cite{hasinoff2016burst} \cite{yu2016lsun}, celebrity faces \cite{liu2015faceattributes}, or 3D human hair \cite{uschairsalon}. However, there are no datasets created for fur on more primitive geometries. Authoring and rendering fur with ray tracing to achieve photorealistic quality is a time consuming process; as such, our contribution can help save the time and expertise required to create such a dataset.

Neural rendering has many applications, ranging from denoising \cite{jaindenoise}, supersampling \cite{fbneuralsupersampling} \cite{watson2020deep}, neural radiance field \cite{mildenhall2020nerf} \cite{martinbrualla2020nerfw}, volumetric cloud rendering \cite{deepscattering}, to hair synthesis \cite{chai2016} \cite{saito2018} \cite{michigan} and rendering \cite{neuralhair}.
\section{Method}

\subsection{Architecture}
\begin{figure*}[htb]
    \centering
    \includegraphics[width=0.9\textwidth]{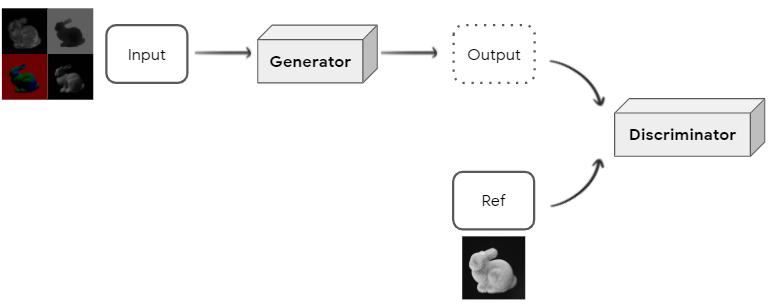}
    \caption{Architecture of the conditional GAN. The inputs are rasterized input buffers, and the ground truth reference is the ray traced image.}
    \label{fig:fur gan architecture}
\end{figure*}

To validate iterations of our dataset, we use a GAN \cite{goodfellow2014generative} image to image model. Our network consists of a convolutional UNet generator $\mathcal{G}$ with skip connections and a convolutional discriminator $\mathcal{D}$ (see fig. \ref{fig:fur gan architecture}). Instance normalization is applied to the convolutional layers in $\mathcal{G}$ and $\mathcal{D}$. We use Adam for optimization. We use WGAN-GradientPenalty loss \cite{wgangulrajani2017}, perceptual loss \cite{johnson2016perceptual} and L1 loss. Our model uses data augmentation with random horizontal flopping and vertical flipping to increase the training sample diversity. 

\subsection{Conditioning inputs}
We use rasterized buffers as conditioning inputs that propagate through $\mathcal{G}$ and $\mathcal{D}$. The input images are generated as 512x512 patches to better fit into memory. The patches are sampled based on the Mask buffer to maximize using samples that contain the region of fur we want the model to observe.

\section{Dataset Preparation}

The dataset consists of several different parts. We provide the 3D groom assets that are used to render the fur, the rendered images of the fur, and a variety of representations that are cheap to calculate from a game engine. Finally, we include the full high quality simulations of fur positions.

\textit{Groom asset}. We use the Houdini software to procedurally generate the groom asset. Starting with a base geometry, or the skin primitive, we distribute guide curves evenly along the skin primitive surface and added small amounts of curliness and lifting so that the fur would stick out.

\textit{Ground truth images}. To render the ground truth images, we used Houdini’s rendering engine Mantra to ray trace the groom asset, producing the high quality ground truth images. Due to the complexity of ray tracing fur, which has a high density of polygons and transparent materials, the process of rendering each frame can take up several minutes, or hours for a full video sequence on a local machine. We opted to use Google Zync Cloud Rendering infrastructure to speed up the process. This has the advantage of distributing the rendering jobs across multiple machines, significantly reducing the time that it would normally take for fur rendering.

\textit{Input buffers images}. To create the various input buffer images, we use the viewport of Houdini to render the scene depth, world normal, lit primitive, mask and guide curves buffers. Because the Houdini viewport is a real-time rasterized rendering engine, our input buffers would be equivalent to captures taken from other similar real-time rendering engines. We use the mask buffer only during this step to select relevant path samples that are more likely to contain fur over uninteresting details like the dark backgrounds.

\textit{Simulation Data}. We use the Alembic format, as it has a widely supported API, to export the simulation data for each scene sequence. We added a custom attribute to specify the number of segments per strand.

\textbf{Note:} To provide more diversity in lighting, we rotate our light rig exactly twice for 360 frames each. Once along the camera's up vector, and once along the camera's forward vector. Therefore, it can be useful to compare results from frame 1, 180, 360, 540).

\section{Results}
\newcommand\figurewidth{.2\linewidth}

\newcommand\imggenerated[2]{
    \begin{subfigure}[b]{\figurewidth}
        \includegraphics[width=\linewidth]{#1}
        \caption{Input}
    \end{subfigure}
    \begin{subfigure}[b]{\figurewidth}
        \includegraphics[width=\linewidth]{#2}
        \caption{Generated}
    \end{subfigure}    
}

\begin{figure*}[thp]
    \centering
    \imggenerated{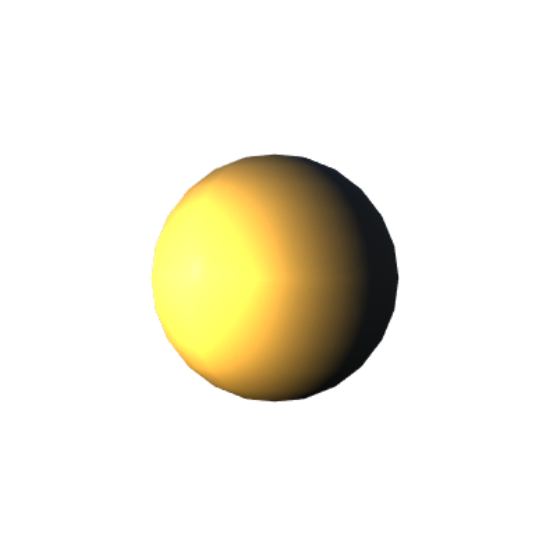}
        {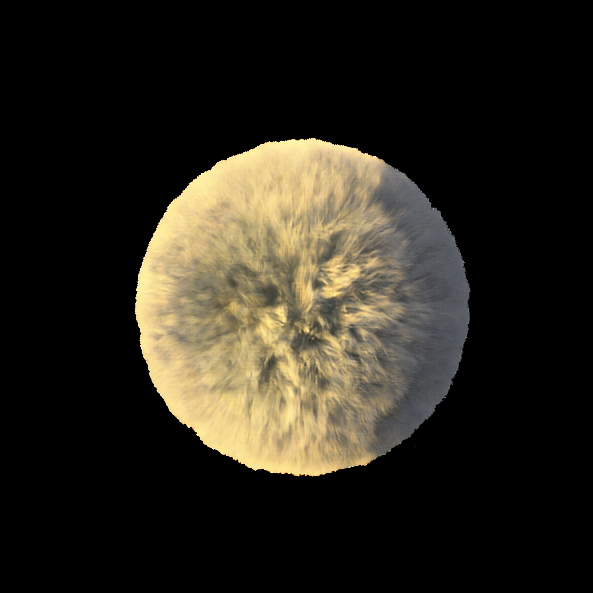}
    \imggenerated{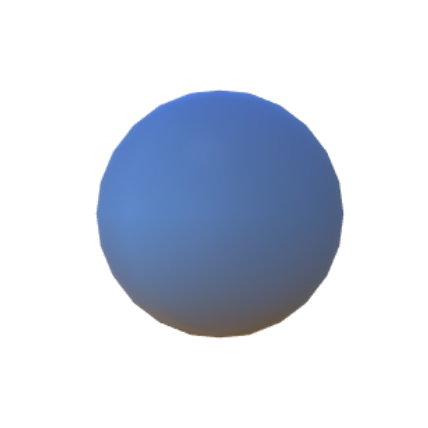}
        {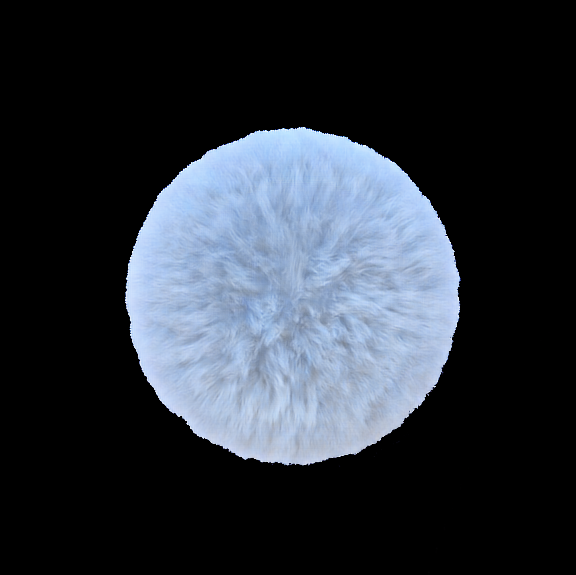}
    \imggenerated{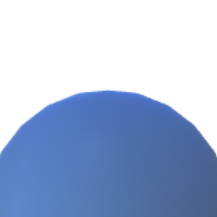}
        {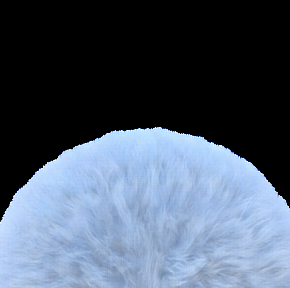}
    \imggenerated{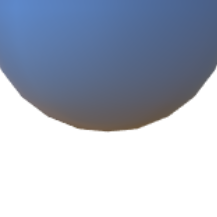}
        {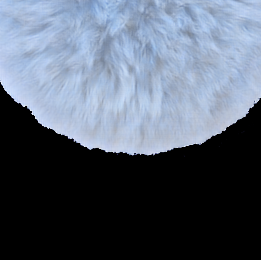}
        
    \caption{By providing different lit primitive shadings in addition to the guide curves, the model can generalize to a different lighting environment. In (e), (f), the lighting is bluer at the top of the sphere, compared to the yellowish color at the bottom of the sphere (g), (h) }
    \label{fig:inf hdri}
\end{figure*}

\begin{figure*}[thp]
    \centering
    \imggenerated{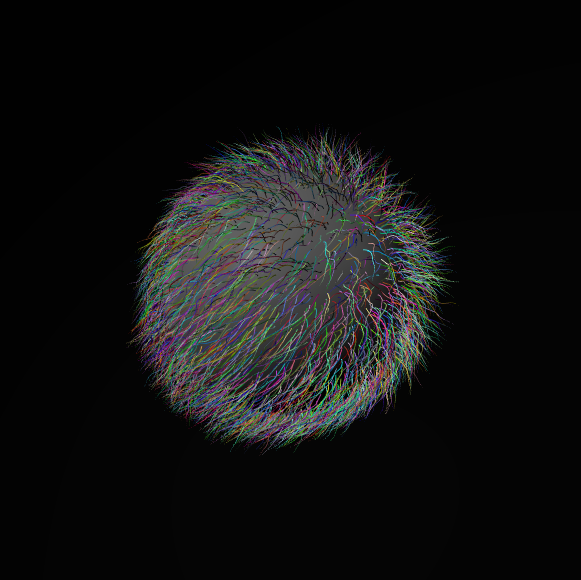}
        {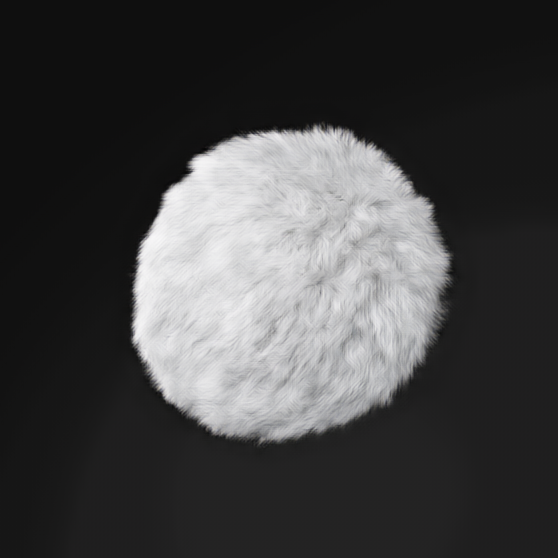}
    \imggenerated{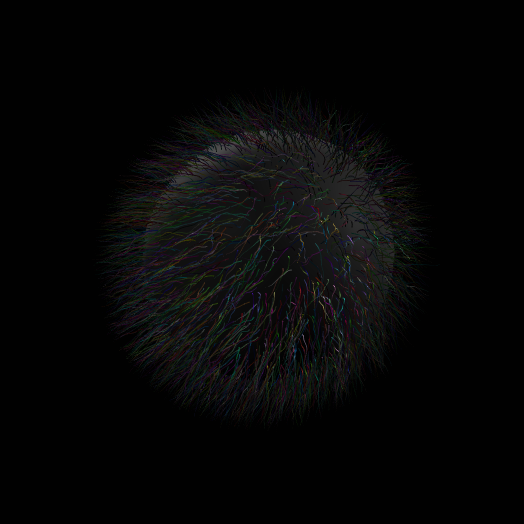}
        {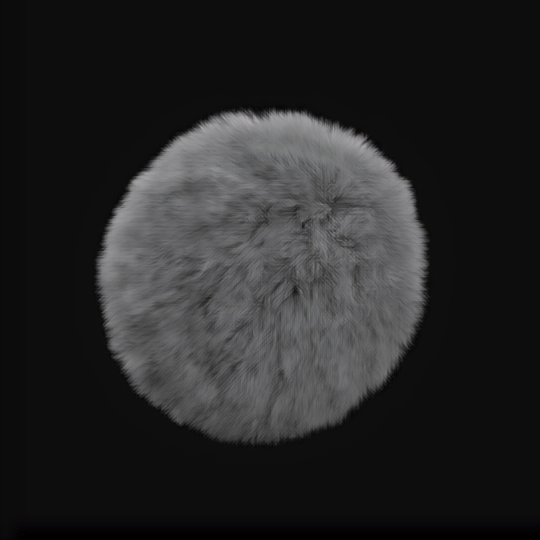}
    \caption{Fur is brighter or darker according to the light intensity.}
    \label{fig:unseen light intensity}
\end{figure*}

\begin{figure*}[thp]
    \centering
    \imggenerated{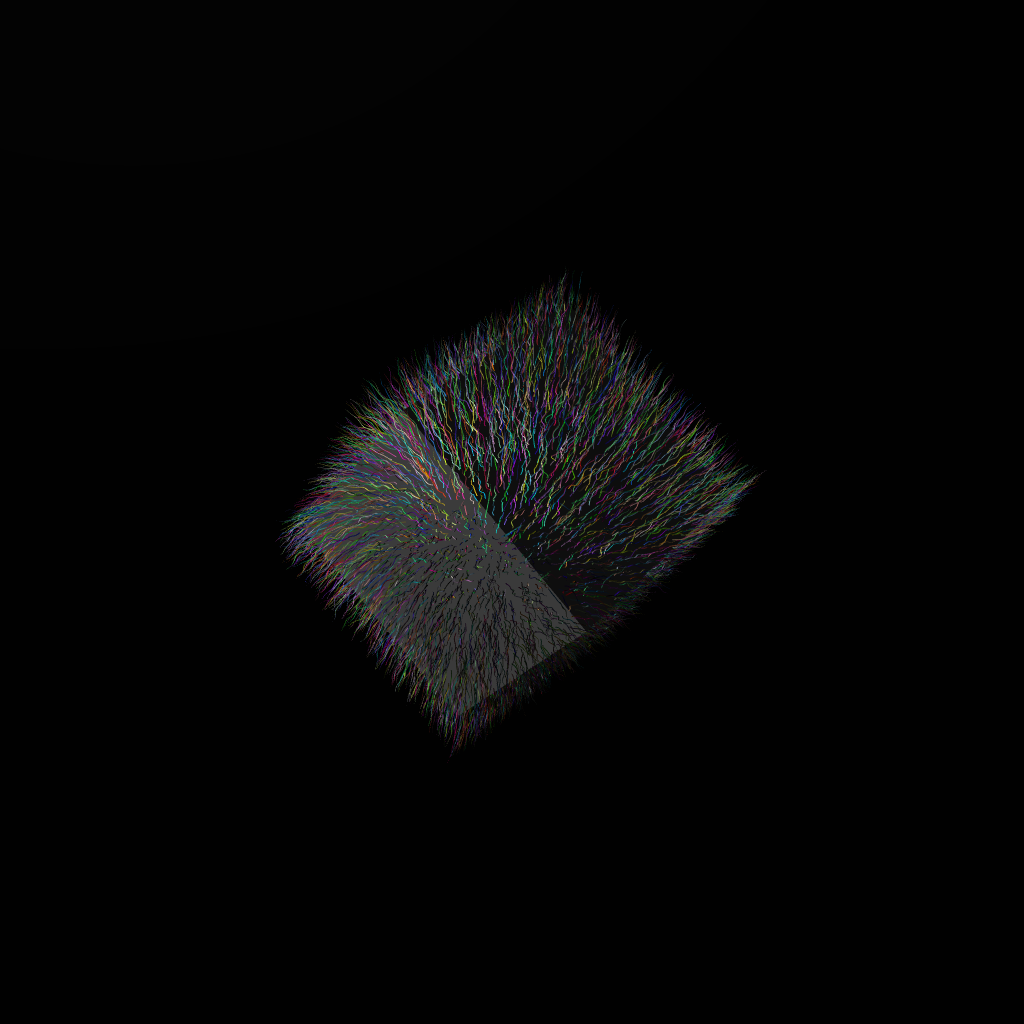}
        {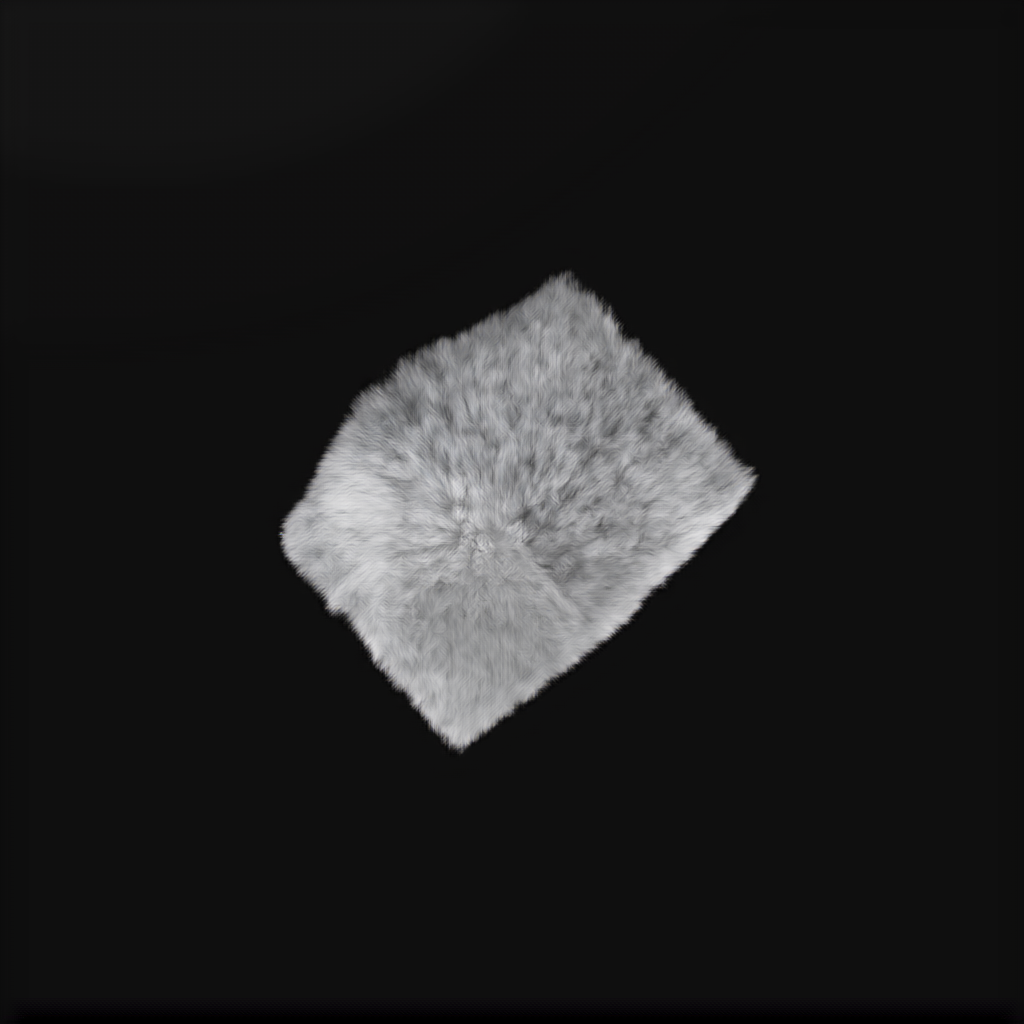}
    \imggenerated{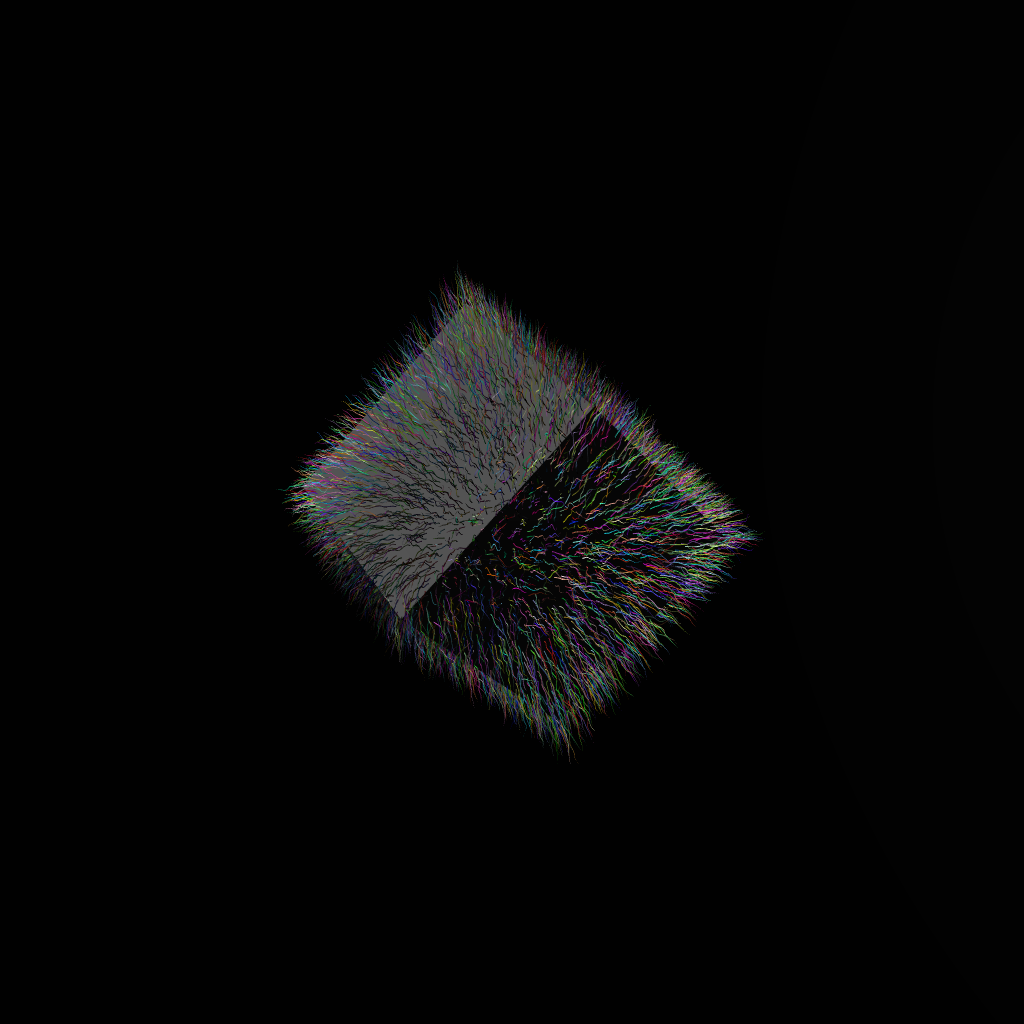}
        {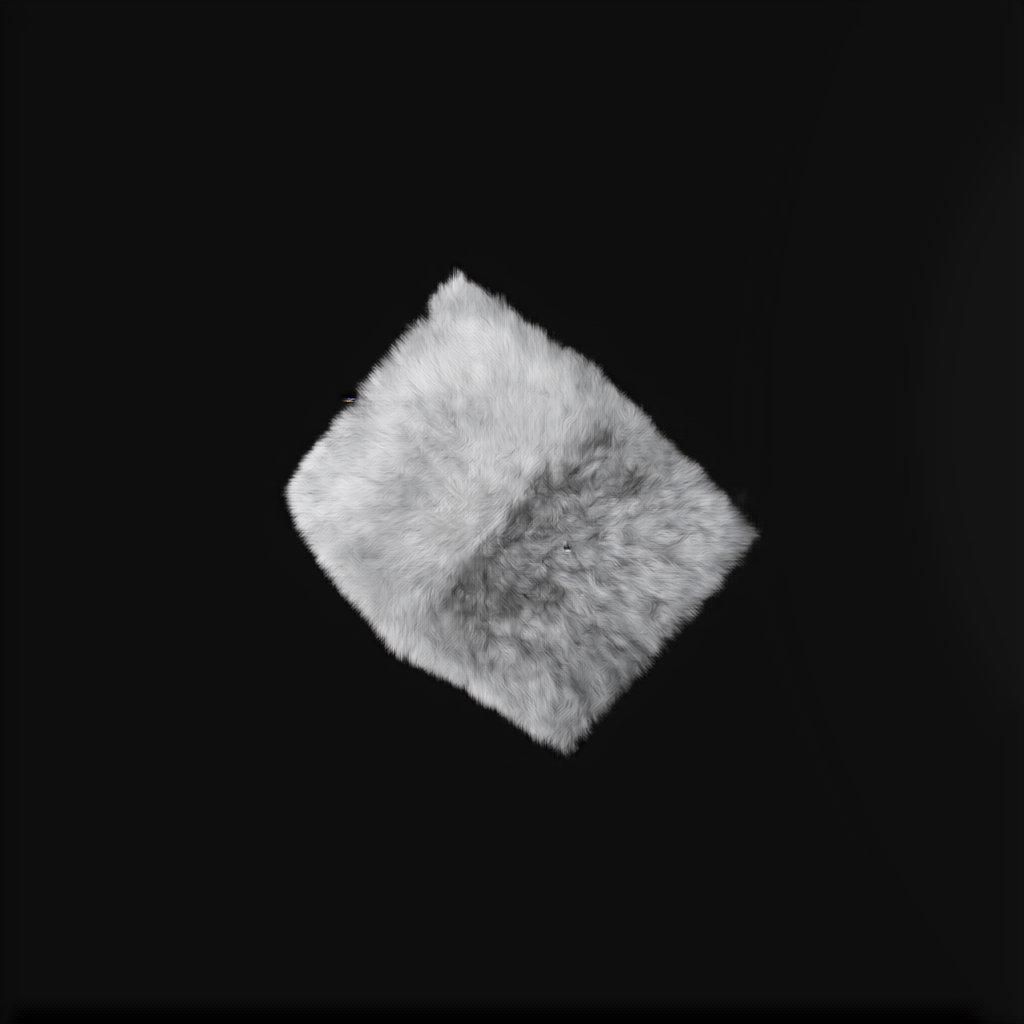}
    \caption{Fur correctly groomed on sharp edges, and shaded the two sides of the boxes correctly based on the diffuse input.}
    \label{fig:unseen sharp edges}
\end{figure*}

\begin{figure*}[thp]
    \centering
    \imggenerated{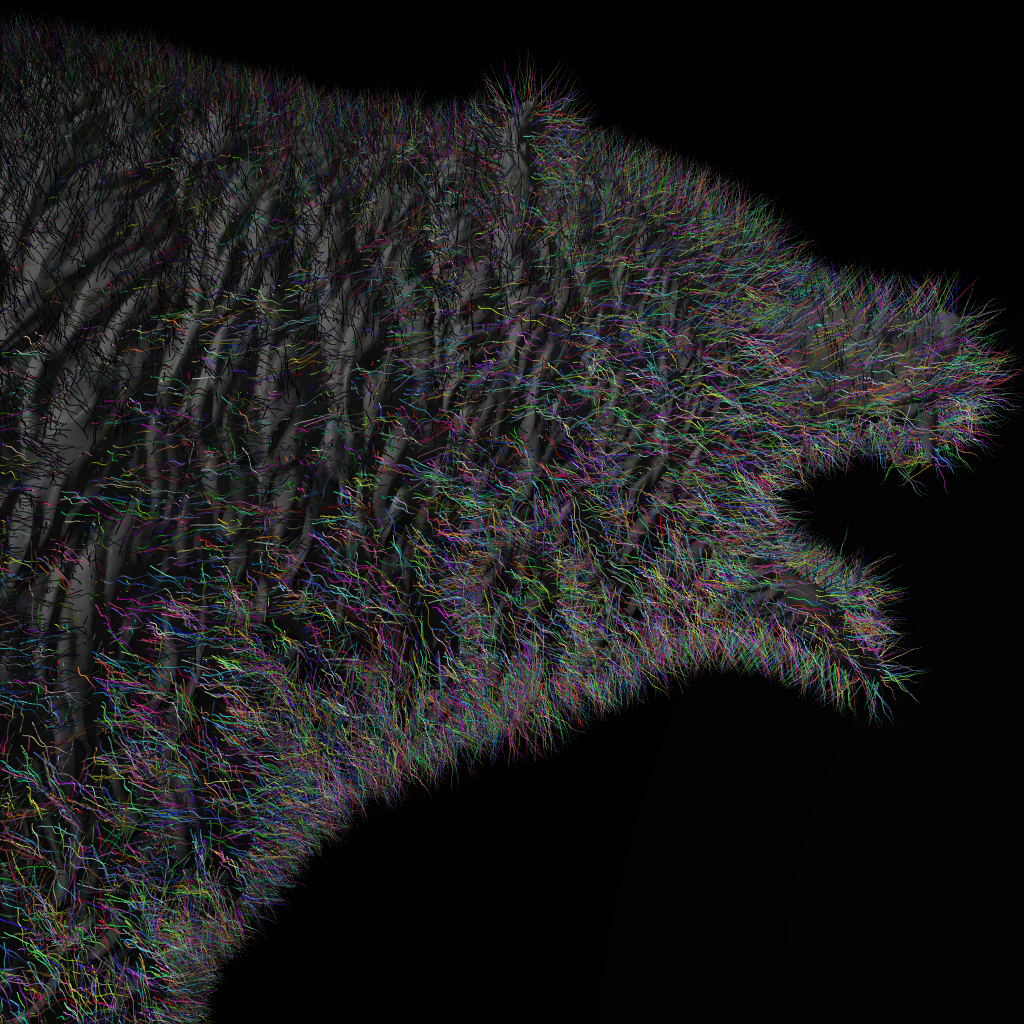}
        {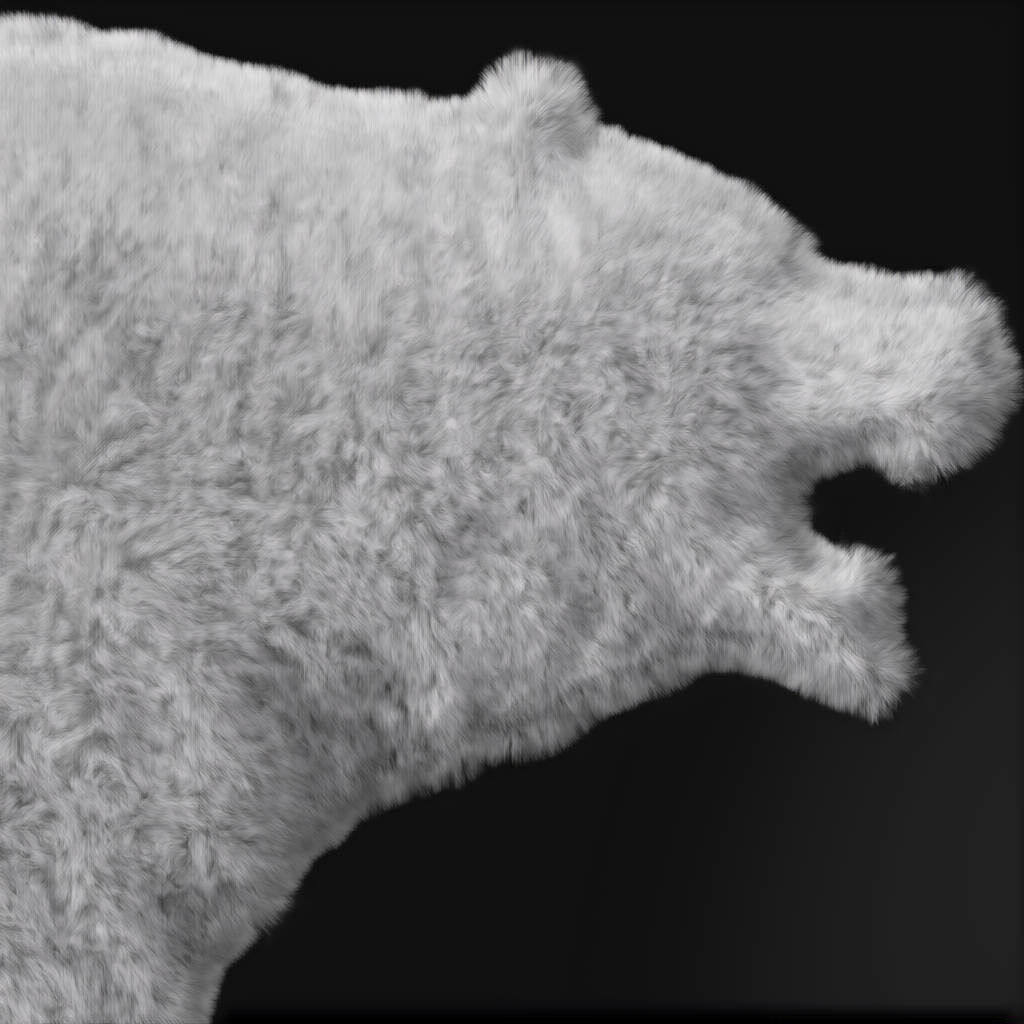}
    \imggenerated{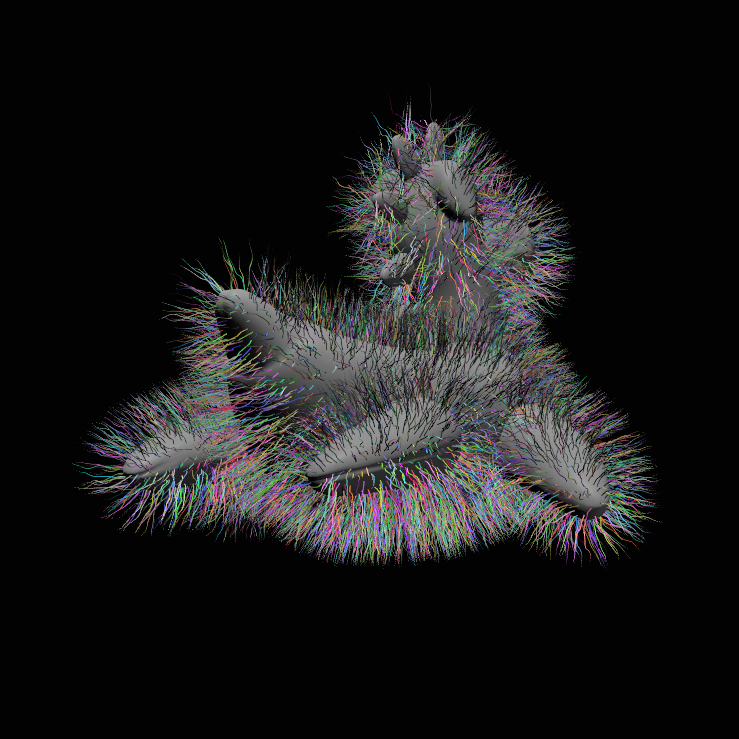}
        {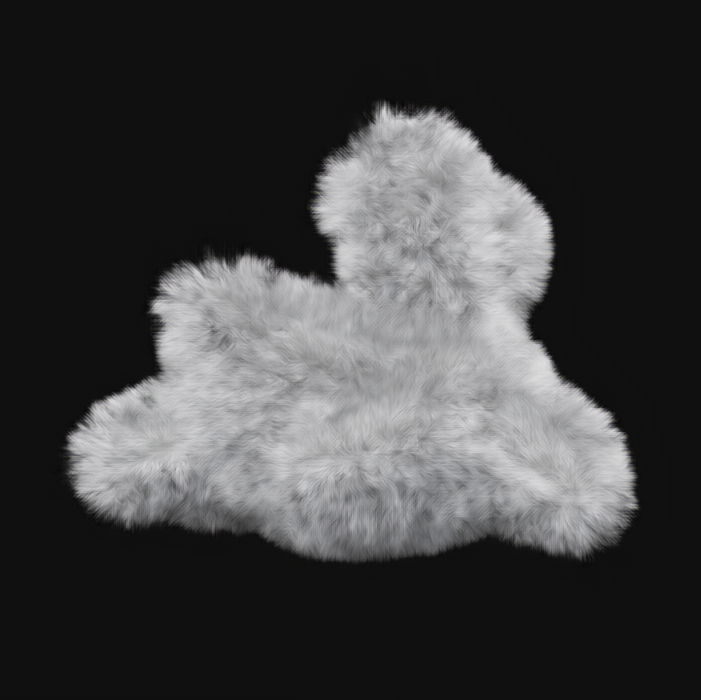}
    \caption{Generalization on more complex skin primitives.}
    \label{fig:unseen complex geometries}
\end{figure*}

\begin{figure*}[thp]
    \begin{subfigure}{\textwidth}
        \includegraphics[width=\textwidth]{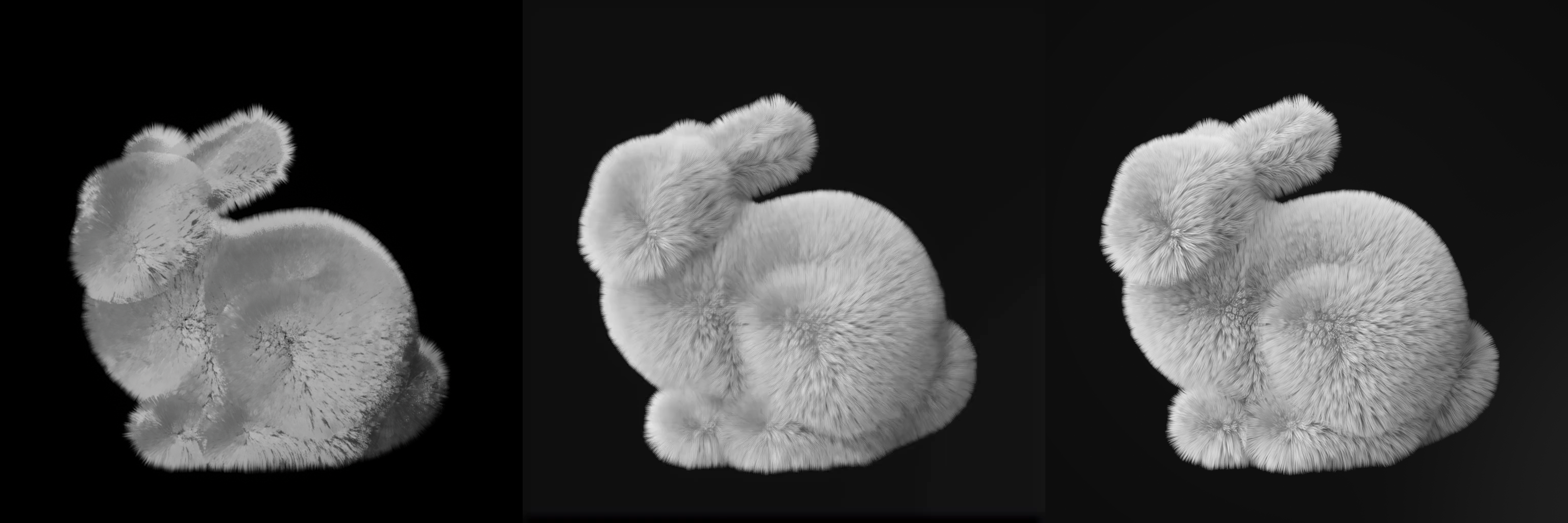}
        \caption{Bunny}
        \label{fig:bunny comparison}
    \end{subfigure}
    \newline
        \begin{subfigure}{\textwidth}
        \includegraphics[width=\textwidth, height=5cm]{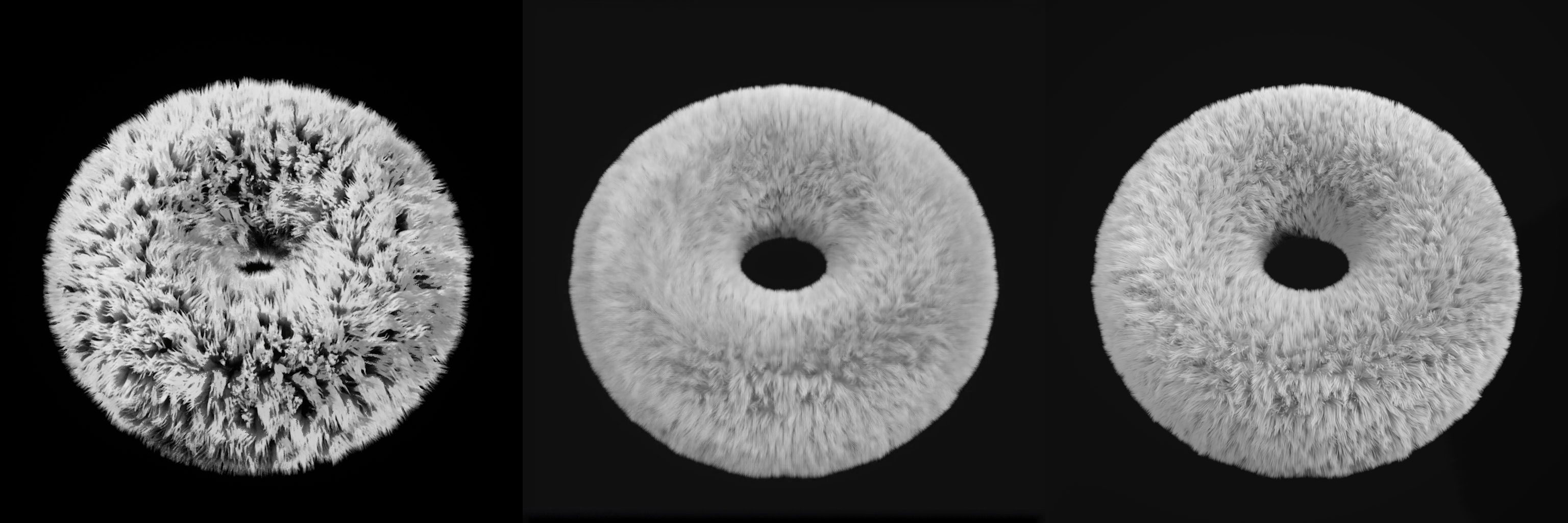}
        \caption{Torus}
        \label{fig:torus comparison}
    \end{subfigure}
    \caption{Comparison of generated images against real-time and offline engines. From left to right: (Real-time) Unreal 4.25, Ours, (Offline) Houdini 17.0.}
    \label{fig:results comparison}
\end{figure*}

\begin{figure*}[thp]
    \centering
    \includegraphics[width=.8\textwidth]{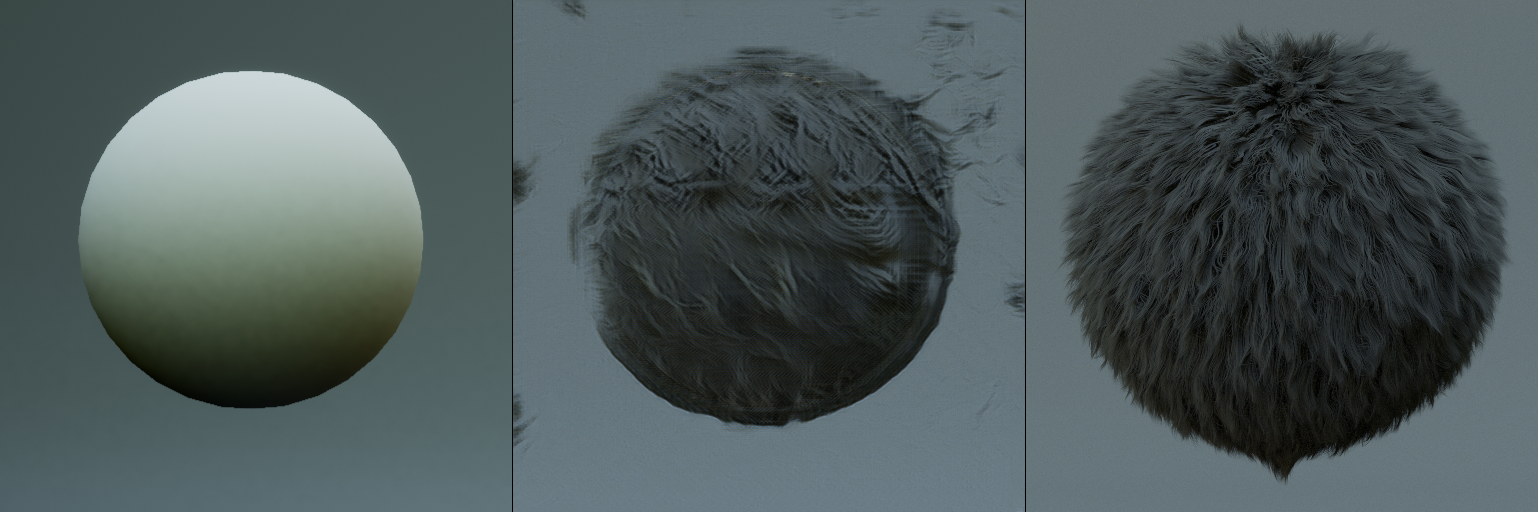}
    \caption{Without the additional guide curve inputs, it is challenging for the model to generate the fur strands properly.}
    \label{fig:no guide curves}
\end{figure*}

\textbf{Note}: All images are generated at resolution 1024 x 1024, but are cropped and resized here to be conveniently displayed. Please see the supplemental materials for the full resolution images.

We train using a combination of different input buffer types. We find that supplying the guide curves, lit primitive, and scene depth inputs provide the best qualitative visual results. The model is able to infer lighting directions from the diffuse shading of the skin primitives, without having to use additional lighting or shadow information. 

We show our results with different lighting environment scenes (fig. \ref{fig:inf hdri}, \ref{fig:unseen light intensity}) and fur generated on unseen skin primitives (fig. \ref{fig:unseen sharp edges}, \ref{fig:unseen complex geometries}).

Using the SyntheticFur dataset, we were able to achieve significantly better visual quality fur rendering than a rasterizer. According to our informal survey of artists, it was significantly closer to offline ray tracing quality (fig. \ref{fig:results comparison}). In particular, we were able to create better soft shadows compared to strand-based real time renderers.
\section{Discussion}

\subsection{Rendering}
We are in the early stage of prototyping the model, with the eventual goal to create an end-to-end pipeline from simulation inputs to rendering outputs for real-time applications. Under the right conditions, our model has shown that it is possible to achieve higher results compared to real-time strand-based approach.

However, to truly be beneficial for real-time applications, we need to improve on performance and add additional control knobs to enable programmers and artists to take advantage of this method. This could be accomplished by using dedicated compute shaders to run inference of a simpler pre-trained model. We would like to implement a modular solution that allows for easy integration into existing workflows of modern rendering engines.

Our approach also has the following additional limitations:
\begin{itemize}
    \item Our generated outputs cannot yet learn proper alpha blending, and therefore will create artifacts during object occlusion. 
    \item While the frames generated are generally stable, more work can be done to further improve temporal consistency. 
    \item Our model is sensitive to gamma changes of input images if they have not been observed during training.
    \item At very sharp grazing angles, fur strands can become very blurry and lose definition.
\end{itemize}

\subsection{Simulation}
We are at the beginning of exploring a neural physics model that use the simulation data to predict fur motion. One approach is to use the hair segment end points to project onto a unit sphere relative to the camera view direction. The projection in this subspace can enable training a 2D CNN more effectively. The points can then be projected back into camera/view space to be used in conjunction with the rendering network. A similar approach to neural physics is demonstrated for cloth \cite{subspaceneuralphysics}.

\section{Acknowledgement}

We would like to thank Omar Skarsvåg and Aaron Parisi for their contributions in creating the dataset, tools to generate training results, and many helpful discussions, and Google Brain and Stadia for enabling this work.

\bibliography{references}

\section{Supplemental Materials}

Ground truth (video): \url{https://youtu.be/5uraQu_5Tyg}.

Inference comparison, Bunny (video): \url{https://youtu.be/_cKmDsjYLbE}.

Inference comparison, Torus (video) : \url{https://youtu.be/rPo8k63OPWI}. 

Dataset website link: \url{https://github.com/google-research-datasets/synthetic-fur}

\subsection{Additional GAN-generated images}

\begin{figure*}[thp]
    \centering
    \includegraphics[width=\textwidth]{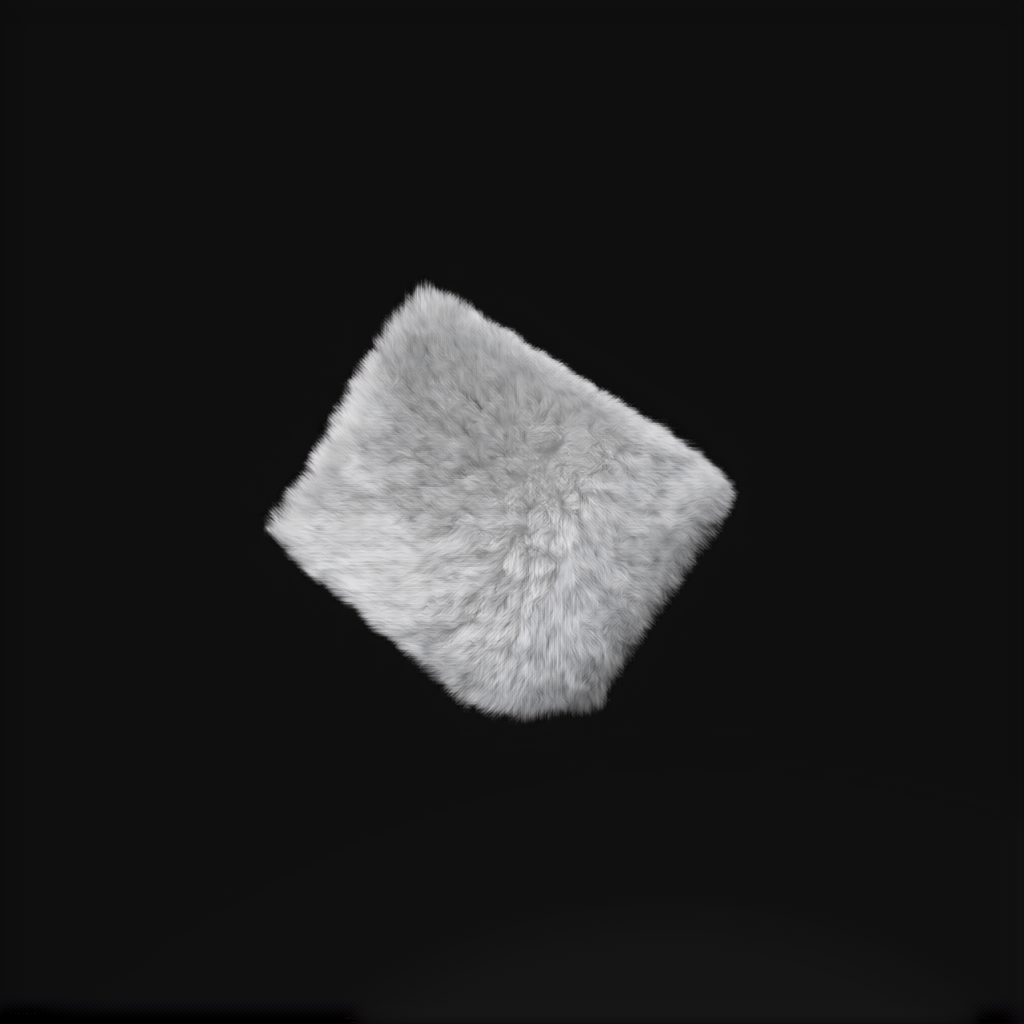}
    \caption{GAN-generated images when trained with SyntheticFur at high resolution.}
    \label{fig:inference box}
\end{figure*}

\begin{figure*}[thp]
    \centering
    \includegraphics[width=\textwidth]{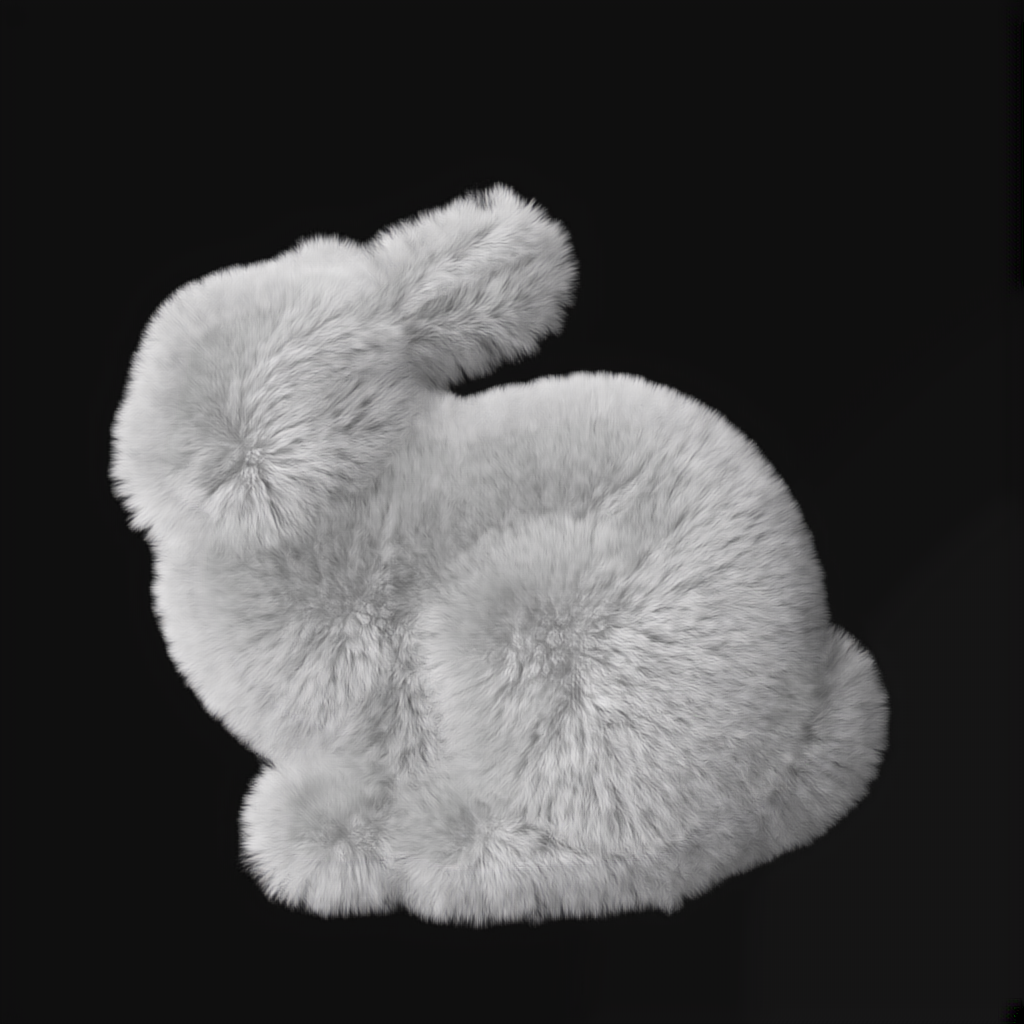}
    \caption{GAN-generated images when trained with SyntheticFur at high resolution.}
    \label{fig:inference bunny}
\end{figure*}

\begin{figure*}[thp]
    \centering
    \includegraphics[width=\textwidth]{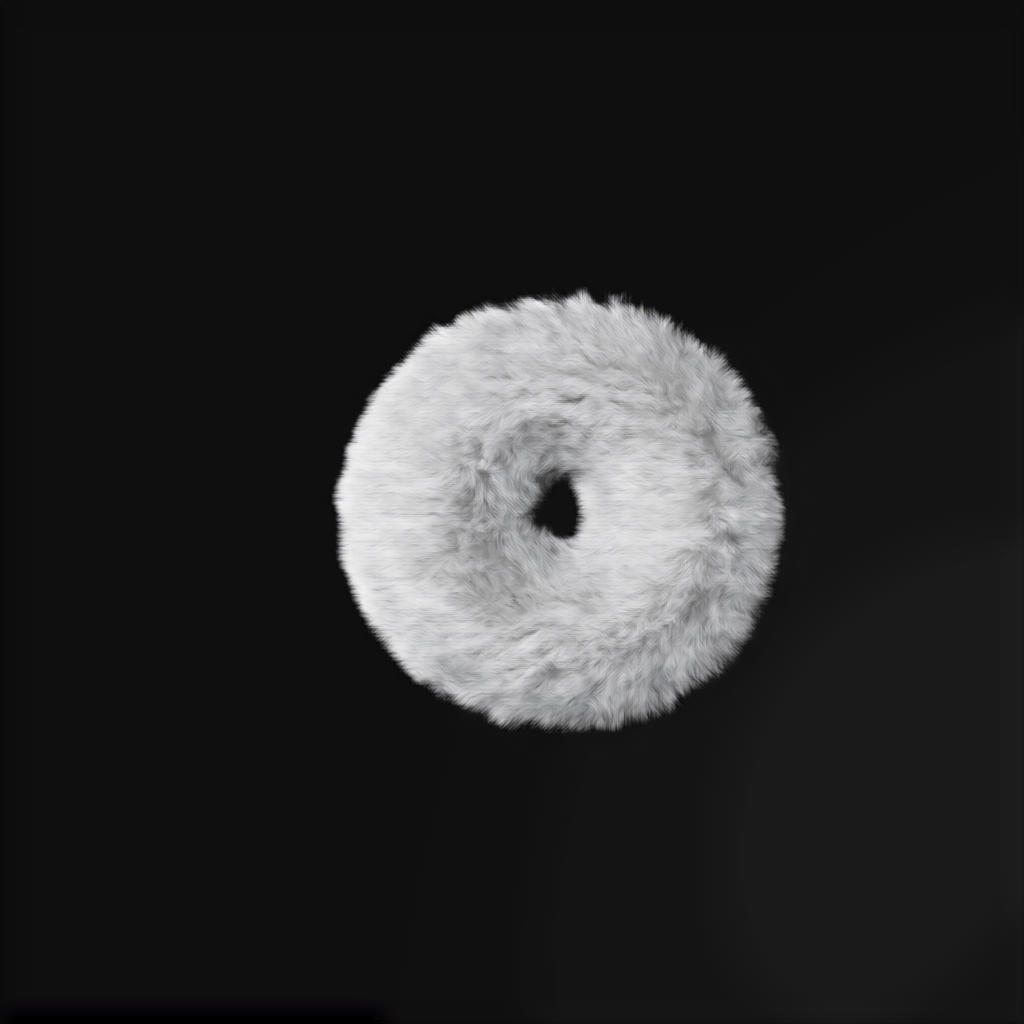}
    \caption{GAN-generated images when trained with SyntheticFur at high resolution.}
    \label{fig:inference torus 1}
\end{figure*}

\begin{figure*}[thp]
    \centering
    \includegraphics[width=\textwidth]{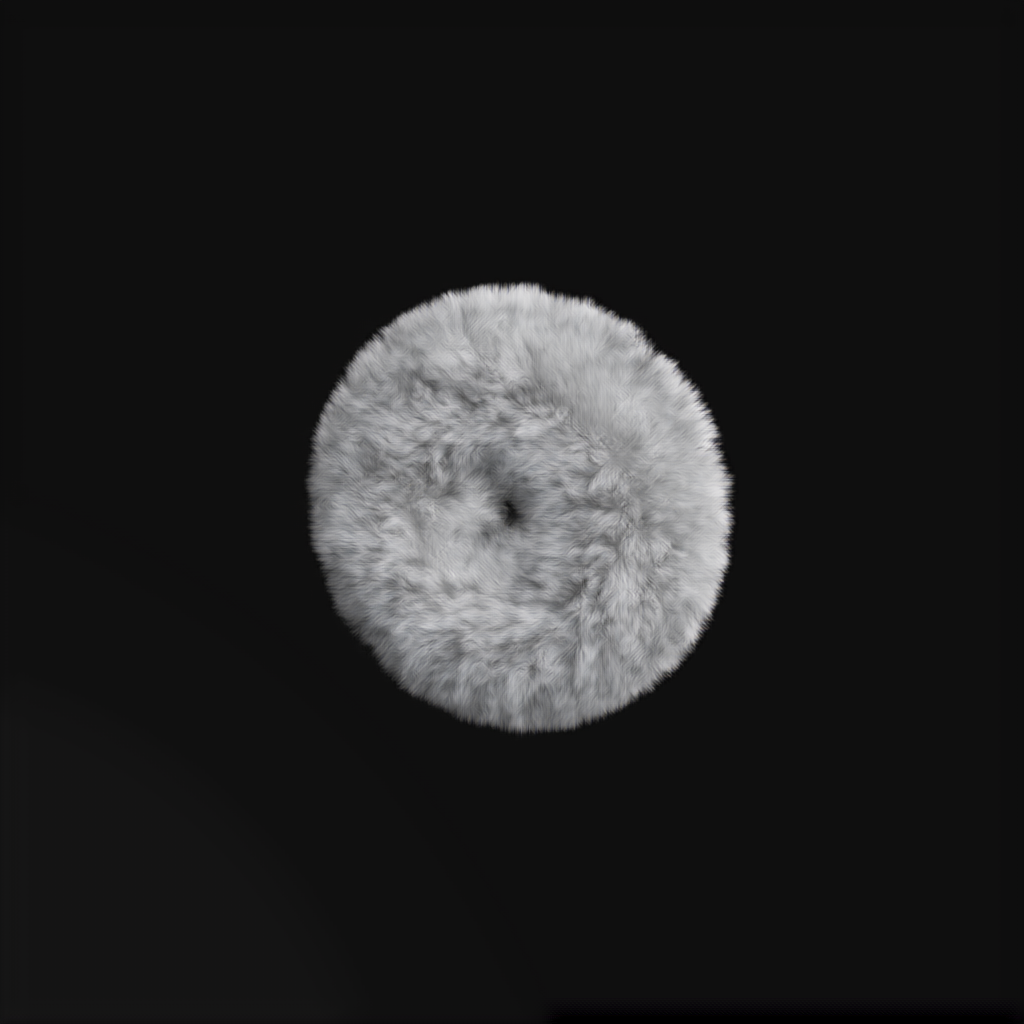}
    \caption{GAN-generated images when trained with SyntheticFur at high resolution.}
    \label{fig:inference torus 2}
\end{figure*}

\begin{figure*}[thp]
    \centering
    \includegraphics[width=\textwidth]{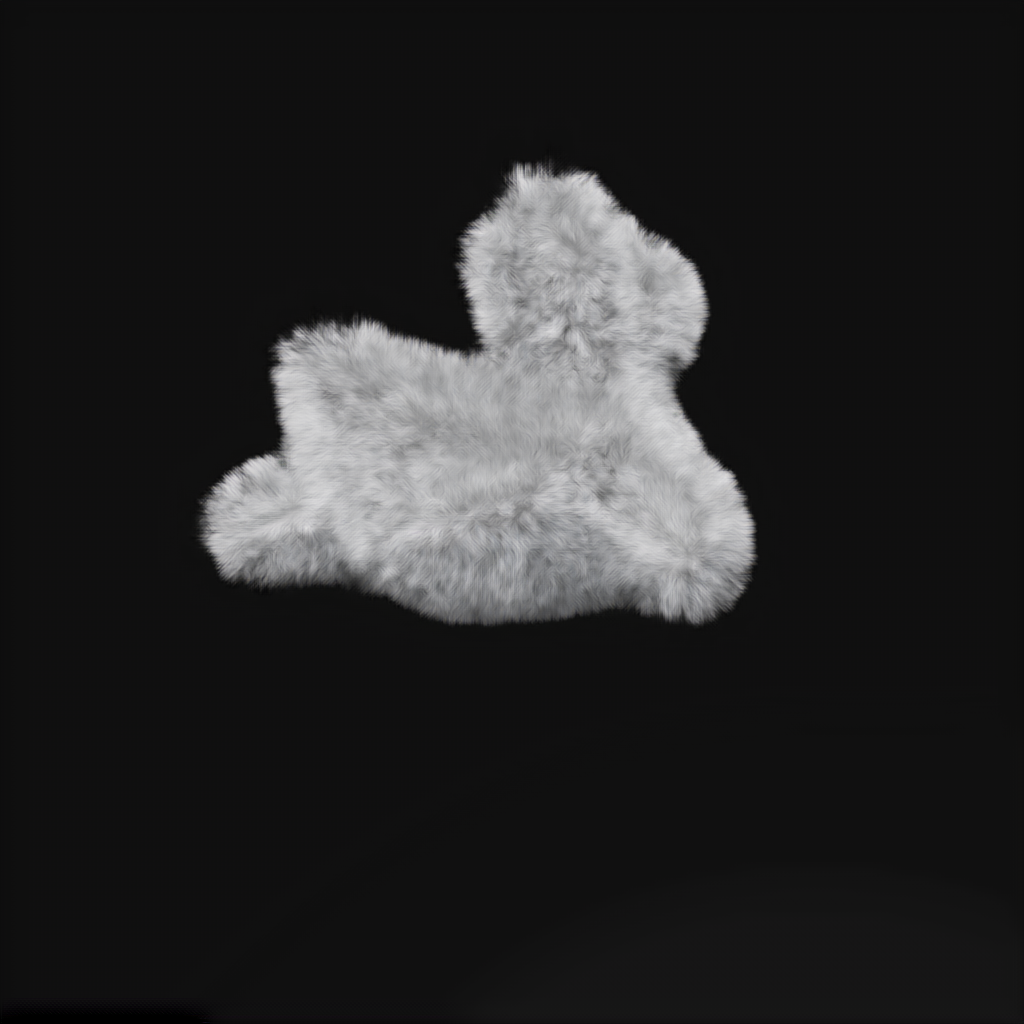}
    \caption{GAN-generated images when trained with SyntheticFur at high resolution.}
    \label{fig:inference rubbertoy}
\end{figure*}

\begin{figure*}[thp]
    \centering
    \includegraphics[width=\textwidth]{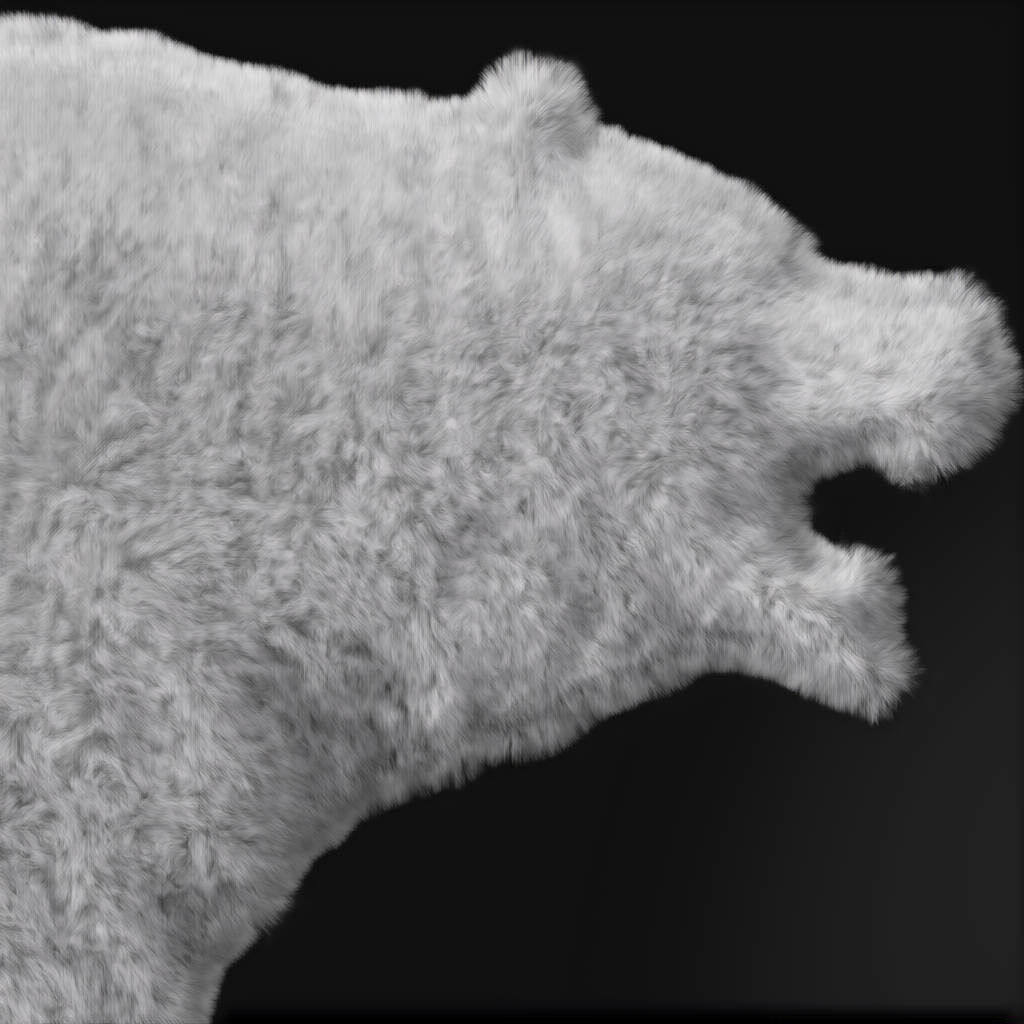}
    \caption{GAN-generated images when trained with SyntheticFur at high resolution.}
    \label{fig:inference bear}
\end{figure*}

\end{document}